\def\year{2019}\relax
\documentclass[letterpaper]{article} 
\usepackage{aaai19}  
\usepackage{times}  
\usepackage{helvet}  
\usepackage{courier}  
\usepackage{url}  
\usepackage{graphicx}  

\usepackage{amsmath,amssymb,amsfonts}

\usepackage[ruled]{algorithm}
\usepackage{algorithmic}
\usepackage{epsfig,epstopdf}
\usepackage{subfigure}
\usepackage{xcolor}

\usepackage{multirow}
\usepackage{array}
\usepackage{booktabs} %

\begin{document}
%
\title{Multiple Independent Subspace Clusterings}
\author{Xing Wang$^1$, Jun Wang$^1$\thanks{Corresponding author, kingjun@swu.edu.cn (Jun Wang)}, Carlotta Domeniconi$^2$, Guoxian Yu$^{1,3}$,  Guoqiang Xiao$^1$, Maozu Guo$^4$\\
$^1$College of Computer and Information Sciences, Southwest University, Chongqing, China\\
$^2$Department of Computer Science, George Mason University, Fairfax, USA\\
$^3$Hubei Key Laboratory of Intelligent Geo-Information Processing, China University of Geosciences, Hubei, China\\
$^4$School of Electrical and Information Engineering, Beijing University Of Civil Engineering and Architecture, Beijing, China\\
Email: \{wx1993cs,kingjun,gxyu,gqxiao\}@swu.edu.cn, carlotta@cs.gmu.edu, guomaozu@bucea.edu.cn\\
}



\maketitle
\begin{abstract}
\begin{quote}
Multiple clustering aims at discovering  diverse ways of organizing  data into clusters. Despite the progress made, it's still a challenge for users to analyze and understand the distinctive structure of each output clustering. To ease this process, we consider diverse clusterings embedded in different subspaces, and  analyze the embedding subspaces to shed light into the structure of each clustering. To this end, we provide a two-stage approach called MISC (Multiple Independent Subspace Clusterings). In the first stage, MISC uses independent subspace analysis to seek multiple and statistical independent (i.e. non-redundant) subspaces, and determines the number of subspaces via the minimum description length  principle. In the second stage, to account for the intrinsic geometric structure of samples embedded in each subspace, MISC performs graph regularized semi-nonnegative matrix factorization to explore clusters. It additionally integrates the kernel trick into matrix factorization to handle non-linearly separable clusters. Experimental results on synthetic datasets show that MISC can find different interesting clusterings from the sought independent subspaces, and it also outperforms other related and competitive approaches on real-world datasets.
\end{quote}
\end{abstract}

\section{Introduction}
Clustering is an unsupervised learning technique that aims at partitioning data into a number of homologous groups (or clusters). However, traditional clustering methods typically provide a single clustering,  and fail to reveal the diverse patterns underlying the data. In fact, several different clustering solutions may co-exist in a given problem, and each may provide a reasonable organization of the data, e.g., people can be assigned to different communities based on different roles;  proteins can be categorized differently based on their amino acid sequences or their 3D structure. In these scenarios, it would be desirable to present multiple alternative clusterings to the users, as these alternative clusterings can explain the underlying structure of the data from different viewpoints.

To address the aforementioned problem, the research field of multi-clustering has emerged during the last decade. Naive solutions run a single clustering algorithm with different parameter values, or explore different clustering algorithms \cite{bailey2013alternative}. These approaches may generate multiple clusterings with high redundancy, since they do not take into account the already explored clusterings. To overcome this drawback, two general strategies have been introduced. The first one  simultaneously generates multiple clusterings, which are required to be different from each other \cite{jain2008simultaneous,dang2010generation}. The second one generates multiple clusterings in a greedy manner, and forces the new clusterings to be different from the already generated ones \cite{cui2007non,hu2015finding,yang2017non}.

Most of these multi-clustering methods consider multiple clusterings in the full feature space. However, as the dimensionality of the data increases, clustering methods encounter the challenge of the {\it {curse of dimensionality}} \cite{parsons2004subspace}. Furthermore, some features may be  relevant to some clusterings but not others. This phenomenon is also observed in data with moderate dimensionality. Subspace clustering aims at finding clusters in subspaces of the original feature space,
but it faces an exponential ($2^{d}-1$) search space and focuses on exploring only one clustering. Some  approaches try to find alternative clusterings in a weighted feature space \cite{caruana2006meta,hu2015finding} or in a transformed feature space \cite{cui2007non,davidson2008finding}; however, the former methods cannot control well the redundancy between different clusterings, and the latter  cannot find multiple orthogonal subspaces at the same time.

To overcome these issues, we propose an approach called Multiple Independent Subspace Clusterings (MISC) to explore diverse clusterings in multiple independent subspaces, one clustering for each subspace. During the first stage, MISC uses  Independent Subspace Analysis (ISA) \cite{szabo2012separation} to explore multiple pairwise-independent (i.e., non-redundant) subspaces by minimizing the mutual information among them, and seeks the number of independent subspaces via  the minimum description length principle \cite{Rissanen2007Information}. MISC automatically determines the number of clusters in each subspace via Bayesian $k$-means \cite{Welling2006Bayesian}, and groups the data embedded in each subspace  using graph regularized semi-nonnegative matrix factorization \cite{ding2010convex}. To group non-linearly separable data in a subspace, it further maps the data into a reproducing kernel Hilbert space via the kernel trick.

This paper makes the following contributions:
\begin{itemize}
\item We introduce an approach called MISC to explore multiple clusterings in independent subspaces. MISC automatically computes the number of independent subspaces, which provide multiple individual views of the data.
\item MISC leverages graph regularized semi-nonnegative matrix factorization and kernel mapping to group non-linearly separable clusters, and can determine the number of clusters in each subspace.
\item Experimental results show that MISC can explore different clusterings in various subspaces, and it significantly outperforms  other related and competitive approaches \cite{caruana2006meta,bae2006coala,cui2007non,davidson2008finding,jain2008simultaneous,hu2015finding,yang2017non,Niu2010Multiple,guan2010variational,niu2012nonparametric}.
\end{itemize}

\section{Related Work}\label{sec:relwork}
Existing multi-clustering approaches can be  classified into two categories depending on how they control redundancy, either based on  clustering labels, or  on  feature space.

COALA (Constrained Orthogonal Average Link Algorithm) \cite{bae2006coala} is the classic algorithm that controls redundancy through clustering labels. It transforms linked pairs of the reference clustering into cannot-link constraints, and then uses agglomerative clustering to find an alternative clustering. MNMF (Multiple clustering by Nonnegative Matrix Factorization) \cite{yang2017non} derives a diversity regularization term from the  labels of existing clusterings, and then integrates this term with the objective function of NMF to seek another clustering. The performance of both COALA and MNMF heavily depends on the quality of already discovered clusterings. To alleviate this issue,  other methods simultaneously seek multiple clusterings by  minimizing the correlation between the labels of two distinct clusterings  and by optimizing the quality of each clustering \cite{jain2008simultaneous,wang2010Muticc}. For example, De-$k$means (Decorrelated $k$-means) \cite{jain2008simultaneous} simultaneously learns two disparate clusterings  by minimizing a $k$-means sum squared
error objective for the two clustering solutions, and by minimizing the correlation between the two clusterings. CAMI (Clustering for Alternatives with Mutual Information) \cite{dang2010generation} optimizes a dual-objective function, in which the log-likelihood objective (accounting for the quality) is maximized, while the mutual information objective (accounting for the dissimilarity) of pairwise clusterings is minimized.

Multi-clustering solutions that explore multiple clusterings using a feature-based criterion have also been studied. Some of them assign weights to  features. For example, MetaC (Meta Clustering) \cite{caruana2006meta} first applies $k$-means to generate a large number of base clusterings using  weighted features based on the Zipf distribution \cite{zipf1949human}, and then obtains multiple clusterings via a hierarchical clustering ensemble. MSC (Multiple Stable Clusterings) \cite{hu2015finding} detects multiple stable clusterings in each weighted feature space using the idea of clustering stability based on Laplacian Eigengap. Unfortunately, MSC cannot guarantee  diversity among multiple clusterings, since it cannot control the redundancy very well. Other feature-wise multi-clusterings are based on  transformed features. They use a data space $S$ to characterize the existing clusterings and try to construct a new feature space, which is either orthogonal to $S$, or independent from $S$. Once the novel feature space is constructed, any clustering algorithm can be used in this space to generate an alternative clustering. OSC (Orthogonal subspace clustering) \cite{cui2007non} transforms the original feature space into an orthogonal subspace using a projection framework based on the given clustering, and then groups the transformed data into different clusters. ADFT (Alternative Distance Function Transformation) \cite{davidson2008finding} adopts a distance metric learning technique \cite{xing2003distance} and singular value decomposition to obtain an alternative orthogonal subspace based on a given clustering. Thereafter, it obtains an alternative clustering by running the clustering algorithm in the new orthogonal feature space. mSC (Multiple Spectral Clusterings) \cite{Niu2010Multiple} finds multiple clusterings by augmenting a spectral clustering objective function, and by using the Hilbert-Schmidt independence criterion (HSIC) \cite{Gretton2005Measuring} among multiple views to control the redundancy. NBMC (Nonparametric Bayesian Multiple Clustering) \cite{guan2010variational} and NBMC-OFV (Nonparametric Bayesian model for Multiple Clustering with Overlapping Feature Views) \cite{niu2012nonparametric} both employ a Bayesian model to explore multiple feature views and clusterings therein.

Feature-based multiple clustering methods typically seek a full space transformation matrix, or measure the similarity between samples in the full space. Therefore, their performance may be compromised with high-dimensional data. Furthermore, some data only show cluster structure on a subset of features. Given the above analysis, we advocate to separately explore diverse clusterings in  independent subspaces, and introduce an approach called MISC. MISC first uses independent subspace analysis to obtain multiple independent subspaces, and then performs clustering in each independent subspace to achieve multiple clusterings. Extensive experimental results show that MISC can effectively uncover multiple diverse clusterings in each identified subspace.

\section{Proposed Approach}\label{sec:method}
MISC  consists of two phases: (1) Finding multiple independence subspaces, and (2) Exploring a clustering in each subspace. In the following, we provide the details of each phase.

\subsection{Independent Subspace Analysis}\label{sec:ISA}
Blind Source Separation (BSS) is a classic problem in signal processing. Independent Component Analysis (ICA) is a statistical technique that can solve the BBS problem  by decomposing complex data into independent subparts \cite{hyvarinen2001ICA}. Let's consider a data matrix $\mathbf{X}\in \mathbb{R}^{d\times n}$ for $n$ samples with $d$ features. ICA describes $\mathbf{X}$ as a linear mixture of sources, i.e., $\mathbf{AS}= \mathbf{X} \in \mathbb{R}^{d\times n}$, where $\mathbf{A}\in \mathbb{R}^{d\times d}$ is the mixing matrix and $\mathbf{S}$ corresponds to the source components. The source matrix $\mathbf{S} \in \mathbb{R}^{d\times n}$  represents $n$ observations under multiple independent row vectors, i.e., $\mathbf{S}=(S_1; S_2; \cdots; S_d)$, where each $S_{i}$ corresponds to a source component.

Unlike ICA, which requires pairwise independence between all individual source components, Independence Subspace Analysis (ISA) aims at finding a linear transformation of the given data, and it yields several jointly independent source subspaces, each of which contains one or more source components. Let's assume there are  $v$ independent subspaces; ISA seeks the corresponding source subspaces $\mathbf{S}^{(1)},\cdots, \mathbf{S}^{(v)}$ by minimizing the mutual information between pairwise subspaces as follows:
\begin{equation}
\begin{aligned}
  \min \textrm{MI}{(\mathbf{S}^{(1)},\cdots, \mathbf{S}^{(v)})}
\end{aligned}
\label{Eq1}
\end{equation}

Various ISA solvers are available, and they vary in terms of the applied cost functions and optimization techniques \cite{szabo2012separation}. For example,  fastISA\cite{Hyv2006FastISA} seeks the mixing matrix $\mathbf{A}$ by iteratively updating its rows in a fixed-point manner. Unfortunately, fastISA can only find equal-sized subspaces, while multiple clusterings may exist in subspaces of different sizes.  Here we adopt a variant of ISA \cite{szabo2012separation}, which makes use of the ``ISA separation principle'', stating that ISA can be solved by first performing ICA, and then searching and merging the components. As such, the independence between the groups is maximized, and the groups do not need to have an equal number of components. This ISA solution only needs to specify the number of subspaces $v$, which is difficult to determine. To compute the number of subspaces, we use a greedy search strategy, which combines  agglomerative clustering and Minimum Description Length (MDL) principle \cite{Rissanen2007Information}.

The first step of agglomerative clustering is to merge subspaces. Given two subspaces $\mathbf{S}^{(i)}$ and $\mathbf{S}^{(j)}$, we compute their independence  as follows:
\begin{equation}
\begin{aligned}
  C_{I}(\mathbf{S}^{(i)},\mathbf{S}^{(j)})=C_{H}(\mathbf{S}^{(i)}\cup\mathbf{S}^{(j)})-C_{H}(\mathbf{S}^{(i)})-C_{H}(\mathbf{S}^{(j)})
\end{aligned}
\label{Eq2}
\end{equation}
where $C_{H}(\mathbf{S})=\frac{|\mathbf{S}|}{2}\cdot\log_{2}(n)+\sum_{i=1}^{n}\log_{2}\frac{1}{f_{\mathbf{S}}(\mathbf{S}_{\cdot i})}$ is the entropy cost to encode the $n$ objects in the subspace $\mathbf{S}$ using the probability-density function $\frac{1}{f_{\mathbf{S}}}$, which can be obtained using kernel density estimation\footnote{https://bitbucket.org/szzoli/ite/downloads/}. We compute $C_{I}$ of each pair of subspaces and merge the subspaces with the smallest $C_{I}$. We repeat the above step until the number of subspaces
$v<2$, or all the $C_{I}>0$.

We apply the MDL principle to determine the number of subspaces. MDL is widely used for model selection. Its core idea is to choose the model, which allows a receiver to exactly reconstruct the original data using the most succinct transmission. MDL balances the coding length of the model and the coding length of the deviations of
the data from that model. More concretely, the coding cost for transmitting data $D$ together with a model $M$ is
\begin{equation}
\begin{aligned}
  L(D,M)=L(M)+L(D|M)
\end{aligned}
\label{Eq3}
\end{equation}

When  subspaces are merged in each iteration, we update  $L(D,M)$.  Finally, we choose the number of subspaces $v$ corresponding to the smallest $L(D,M)$. Concretely, we use the technique in \cite{Rissanen2007Information,Ye2016Generalized} to measure the length of the model  and data coding as follows:
\begin{equation}
\begin{aligned}
  L(M)=\frac{d^2}{2}\cdot\log_{2}(n)+(v+1)\cdot\log_{2}(d)
\end{aligned}
\label{Eq4}
\end{equation}

\begin{equation}
\begin{aligned}
  L(D|M)=\frac{d}{2}\cdot\log_{2}(n)+\sum_{i=1}^{v}\sum_{j}^{n}\log_{2}\frac{1}{f_{\mathbf{S}}(\mathbf{S}_{\cdot j}^{(i)})}
\end{aligned}
\label{Eq5}
\end{equation}
where $n$ is the number of samples, $d$ is the number of features, and $f_{\mathbf{S}}(\mathbf{S}_{\cdot j}^{(i)})$ is the probability-density function for each subspace. As a result, we obtain $v$ independent subspaces.

\subsection{Exploring Multiple Clusterings}
After obtaining multiple independent subspaces, we use Bayesian $k$-means \cite{Welling2006Bayesian}  to guide the computation of the number of clusters in each subspace. Bayesian $k$-means adopts a variational Bayesian framework \cite{Ghahramani1999Variational} to iteratively choose the optimal number of clusters. We then perform Graph regularized Semi-NMF (GSNMF) to cluster data embedded in each subspace. GSNMF  is an improvement upon SNMF by leveraging  the geometric structure of samples to regularize the matrix factorization.

SNMF \cite{ding2010convex} is a variant of the classical NMF \cite{lee1999learning}; it extends the application of traditional NMF from nonnegative inputs to mix-signed inputs. At the same time, it preserves the strong clustering interpretability. The objective function of SNMF can be formulated as follows:
\begin{equation}
\begin{aligned}
  J_{SNMF}=\parallel\mathbf{X-ZH} \parallel^{2}  s.t. \mathbf{H}\geq 0
\end{aligned}
\label{Eq6}
\end{equation}
where
$\mathbf{Z}\in\mathbb{R}^{d\times k}$ can be viewed as the cluster centroids, and $\mathbf{H}\in\mathbb{R}^{k\times n}, \mathbf{H}\geq 0$ is the soft cluster assignment matrix in the latent space. We can transform the soft clusters to hard clusters by clustering the index matrix $\mathbf{H}$.

Inspired by GNMF (Graph regularized Nonnegative Matrix Factorization) \cite{Cai2011Graph}, we make use of the intrinsic geometric structure of samples to guide the factorization of $\mathbf{H}$, and cascade it to $\mathbf{Z}$. As a result, we obtain the following objective function for the graph regularized SNMF (GSNMF):
\begin{equation}
\begin{aligned}
  J_{GSNMF}=\parallel\mathbf{X}-\mathbf{ZH} \parallel^{2}+ \lambda tr(\mathbf{HLH}^{T})
\end{aligned}
\label{Eq7}
\end{equation}
where $tr(\cdot)$ denotes the trace of a matrix, $\lambda\geq 0$ is the regularization parameter; $\mathbf{L}\in\mathbb{R}^{n\times n}$ is the graph Laplacian matrix $\mathbf{L}=\mathbf{D}-\mathbf{P}$, $\mathbf{P}\in\mathbb{R}^{n\times n}$ is the weighted adjacency matrix of the graph \cite{Cai2011Graph}, $\mathbf{D}\in\mathbb{R}^{n\times n}$ is the diagonal degree matrix whose entries are the row sum of $\mathbf{P}$. By minimizing the graph regularized term, we assume that if  $\mathbf{X}_{\cdot j}$ and $\mathbf{X}_{\cdot i}$ are close to each other, then their cluster labels $\mathbf{H}_{\cdot i}$ and $\mathbf{H}_{\cdot j}$ should be close as well.

However, GSNMF, similarly to NMF and SNMF, does not perform well with data that are non-linearly separable in  input space. To avoid this potential issue, we consider mapping the data points onto a Reproducing kernel Hilbert space $\phi(\mathbf{X})$, and reformulate Eq. (\ref{Eq7})
 as follows:
\begin{equation}
\begin{aligned}
  J_{KGSNMF}=\parallel\mathbf{\phi(\mathbf{X})-ZH} \parallel^{2}+ \lambda Tr(\mathbf{HLH}^{T})
\end{aligned}
\label{Eq8}
\end{equation}

This formulation makes it difficult to compute $\mathbf{Z}$ and $\mathbf{H}$, since they depend on the mapping function $\phi(\cdot)$. To solve this problem, we add  constraints on the basis vectors $\mathbf{Z}$. As such, the basis matrix $\mathbf{Z}$ can be further formulated as the combination of weighted-samples $\mathbf{Z}=\phi(\mathbf{X})\mathbf{W}$, in which $\mathbf{W}\geq0$ is the weight matrix. Eq. (\ref{Eq8}) can be rewritten as follows:
\begin{equation}
\begin{aligned}
  J_{KGSNMF}=\parallel\mathbf{\phi(\mathbf{X})-\phi(\mathbf{X})\mathbf{W}H} \parallel^{2}+ \lambda
  Tr(\mathbf{HLH}^{T})\\
  s.t. \mathbf{W}\geq 0; \mathbf{H}\geq 0
\end{aligned}
\label{Eq9}
\end{equation}
Through  kernel mapping,  KGSNMF can properly cluster, not only linearly separable data, but also non-linearly separable ones.

\textbf{Optimization}: We follow the idea of standard NMF to optimize $\mathbf{W}$ and $\mathbf{H}$ by an alternating optimization technique. Particularly, we alternate the optimization of $\mathbf{W}$ and $\mathbf{H}$, while fixing the other  as constant. For simplicity, we use $\phi$ to represent $\phi(\mathbf{X})$.

Optimizing $J_{KGSNMF}$ with respect to $\mathbf{W}$ is equivalent to optimizing the following function:
\begin{equation}
 \begin{aligned}
   J_{1}(\mathbf{W}) = \parallel\mathbf{\phi-\phi\mathbf{W}H} \parallel^{2}
  \end{aligned}
  \label{Eq10}
\end{equation}
To embed the constraint $\mathbf{W}\geq0$, we introduce the Lagrange multiplier $\mathbf{\Phi} \in \mathbb{R}^{n\times k}$:
\begin{equation}
   \begin{aligned}
    L(\mathbf{W}) & = \parallel\mathbf{\phi-\phi\mathbf{W}H} \parallel^{2} - \mathbf{\Phi} \mathbf{W}^{T}
  \end{aligned}
  \label{Eq11}
\end{equation}
Letting the partial derivative $\frac{\partial L(\mathbf{W})}{\partial{\mathbf{W}}}=0$, we obtain
\begin{equation}
   \begin{aligned}
    \mathbf{\Phi} & = (\phi^{T}\phi+\phi^{T}\phi\mathbf{WH})\mathbf{H}^{T}
  \end{aligned}
  \label{Eq12}
\end{equation}
Based on the Karush-Kuhn-Tucker (KKT) \cite{boyd2004convex} complementarity condition $\mathbf{\Phi}_{ij}\mathbf{W}_{ij}=0$, we have:
\begin{equation}
   \begin{aligned}
    {[(\phi^{T}\phi+\phi^{T}\phi\mathbf{WH})\mathbf{H}^{T}]_{ij}}\mathbf{W}_{ij}=0
  \end{aligned}
  \label{Eq13}
\end{equation}
Eq. (\ref{Eq13}) leads to the following updating formula for $\mathbf{W}$:
\begin{equation}
\begin{aligned}
  \mathbf{W}_{ij}\leftarrow \mathbf{W}_{ij}\sqrt{\frac{{[\phi^{T}\phi]^{+}\mathbf{H}^{T}+[\phi^{T}\phi]^{-}\mathbf{W}\mathbf{H}\mathbf{H}^{T}_{ij}}}{{[\phi^{T}\phi]^{-}\mathbf{H}^{T}+[\phi^{T}\phi]^{+}\mathbf{W}\mathbf{H}\mathbf{H}^{T}_{ij}}}}
\end{aligned}
\label{Eq14}
\end{equation}
where we separate the positive and negative parts of $\phi^{T}\phi$ by setting $
[\phi^{T}\phi]^{+}=(|\phi^{T}\phi|+\phi^{T}\phi)/2,[\phi^{T}\phi]^{-}=(|\phi^{T}\phi|-\phi^{T}\phi)/2.$

Similarly, we can get the updating formula for $\mathbf{H}$:
\begin{equation}
\begin{aligned}
  \mathbf{H}_{ij}\leftarrow \mathbf{H}_{ij}\sqrt{\frac{{\mathbf{W}^{T}[\phi^{T}\phi]^{+}+\mathbf{W}^{T}[\phi^{T}\phi]^{+}\mathbf{WH}+\lambda[\mathbf{HL}]^{-}_{ij}}}{{\mathbf{W}^{T}[\phi^{T}\phi]^{+}+\mathbf{W}^{T}[\phi^{T}\phi]^{+}\mathbf{WH}+\lambda[\mathbf{HL}]^{+}_{ij}}}}
\end{aligned}
\label{Eq15}
\end{equation}
From  Eq. (\ref{Eq14}) and Eq. (\ref{Eq15}), we can see that the updating formulas do not depend on the mapping function $\phi(\cdot)$, and we can compute $\phi(\mathbf{X}_{\cdot i})^{T}\phi(\mathbf{X}_{\cdot j})$ via any  kernel function, i.e., $\phi(\mathbf{X}_{\cdot i})^{T}\phi(\mathbf{X}_{\cdot j})=\kappa(\mathbf{X}_{\cdot i},\mathbf{X}_{\cdot j})$.

By iteratively applying Eqs. (\ref{Eq14}) and (\ref{Eq15}) in each independent subspace, we can obtain the optimized $\mathbf{W}^*$ and $\mathbf{H}^*$. Each $\mathbf{H}^{*}$ obtained from each subspace corresponds to one clustering. As such, we obtain $v$ clusterings from $v$ independent subspaces.

Algorithm \ref{alg1} presents the whole MISC procedure.  Line 1 computes the source matrix $\mathbf{S}$ via independent component analysis; Line 2 merges the subspaces according to Eq.(\ref{Eq2}) and using agglomerative clustering, and saves the MDL ($L_{min}$) for each merge; Line 3 chooses the best set of subspaces $\Omega_{min}$ with the minimum MDL ($L_{i}$) through a sorting operation; Lines 5-9 cluster data for each subspace though KGSNMF; Lines 10-18 give the procedure of KGSNMF.

\begin{algorithm}
\caption{MISC: Multiple Independent Subspace Clusterings}
  \begin{algorithmic}[1]
  \REQUIRE ~~$\mathbf{X}$: dataset of $n$ samples with $d$ features;
  \ENSURE ~~$\{{\mathcal{C}_i}\}_{i=1}^v$ : $v$ clusterings.\\
     \STATE $\mathbf{S} =$ ICA$(\mathbf{X})$
     \STATE $\{L_{i},\Omega_{i}\}_{i=1}^{d}$= MergeSubspace($\mathbf{S}$) /*agglomerative clustering $\mathbf{S}$; $\Omega_{i}$ is the set of subspaces after $i$-th mergeing and $L_{i}$ is the MDL corresponding to  $\Omega_{i}$*/
     \STATE $\{L_{min},\Omega_{min}\}$=sort($\{L_{i},\Omega_{i}\}$).\\ /*$\Omega_{min}=\{\mathbf{S}^{(1)}, \mathbf{S}^{(2)}, \cdots, \mathbf{S}^{(v)}\}$*/
     \STATE \textbf{For} $j=1:|\Omega_{min}|$ \quad
     \STATE \quad $k_{j}$ = Bayesian $k$-means($\mathbf{S}^{(j)}$)
     \STATE \quad $\mathbf{H}^{(j)}$ = KGSNMF($\mathbf{S}^{(j)}, k_{j}$)
     \STATE \quad $\mathcal{C}_{j}$ = $k$-means($\mathbf{H}^{(j)}, k_{j}$)
     \STATE \textbf{End For}
     \STATE \textbf{Function} $\mathbf{H}$ = KGSNMF($\mathbf{X}$,$k$)
     \STATE \quad Initialize $\mathbf{W}$ and $\mathbf{H}$ randomly.
     \STATE \quad /* Compute kernel similar matrix*/ \\
     \STATE \quad $[\phi(\mathbf{X}_{\cdot i})^{T}\phi(\mathbf{X}_{\cdot j})] = \kappa(\mathbf{X}_{\cdot i},\mathbf{X}_{\cdot j})$
     \STATE \quad \textbf{While} not converged \textbf{Do}
     \STATE \qquad Update $\mathbf{W}$ using Eq. (\ref{Eq14});
     \STATE \qquad Update $\mathbf{H}$ using Eq. (\ref{Eq15});
     \STATE \quad \textbf{End While}
     \STATE \textbf{End Function}
  \end{algorithmic}
  \label{alg1}
\end{algorithm}


\subsection{Complexity analysis}
The complexity of ISA is $O(d^2n)$ and the complexity of MDL is $O(dn^2)$ (for each merge). Since we need to merge the subspaces for at most $d$ times, the overall time complexity of the first stage is $O(d(d^2n+ dn^2))$. For the second stage, MISC takes $O(n^2d)$ time to construct the $p$-nearest neighbor graph. Assuming the multiplicative updates stop after $t$ iterations and the number of clusters is $k$, then the  cost for KGSNMF is $O(tdkn+ n^2d)$. In summary, the overall time complexity of MISC is $O(dn(d^2+ dn+tk+n))$.

\section{Experiments}
\subsection{Experiments on synthetic data}
We first conduct two types of experiments on synthetic data, the first type of experiments is to prove that MISC can find multiple independent subspaces, and the second type is to prove that our KGSNMT has a better clustering performance than SNMF.

\begin{figure}[h!tbp]
\centering
\subfigure[\scriptsize Subspace1]{\label{fig:sy_s1}
\begin{minipage}[r]{0.11\textwidth}
\centering
  \includegraphics[width=\textwidth]{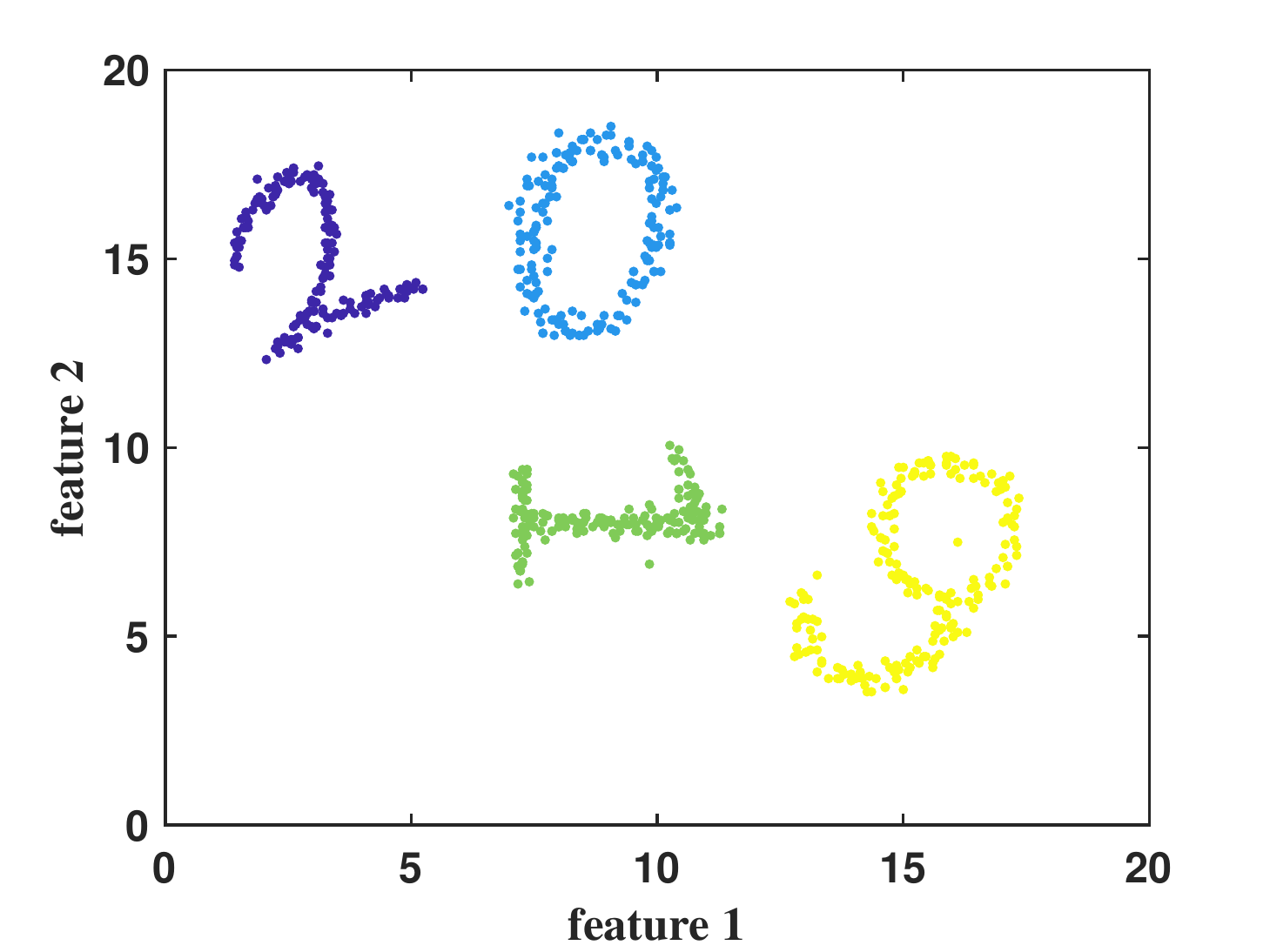}\\
\end{minipage}%
}
\subfigure[\scriptsize Subspace2]{\label{fig:sy_s2}
\begin{minipage}[r]{0.11\textwidth}
\centering
  \includegraphics[width=\textwidth]{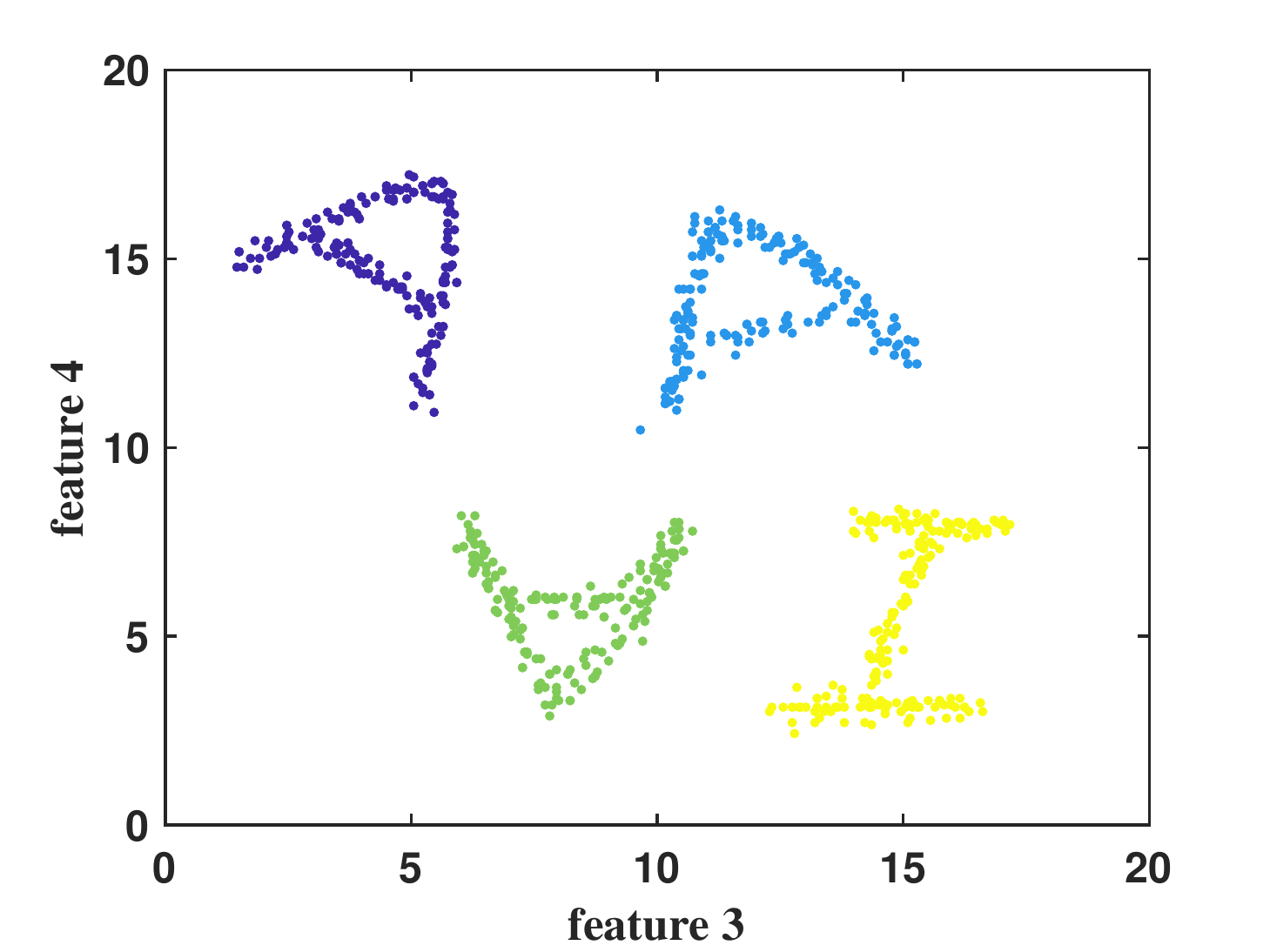}\\
\end{minipage}%
}%
\subfigure[\scriptsize Subspace3]{\label{fig:sy_s3}
\begin{minipage}[r]{0.11\textwidth}
\centering
  \includegraphics[width=\textwidth]{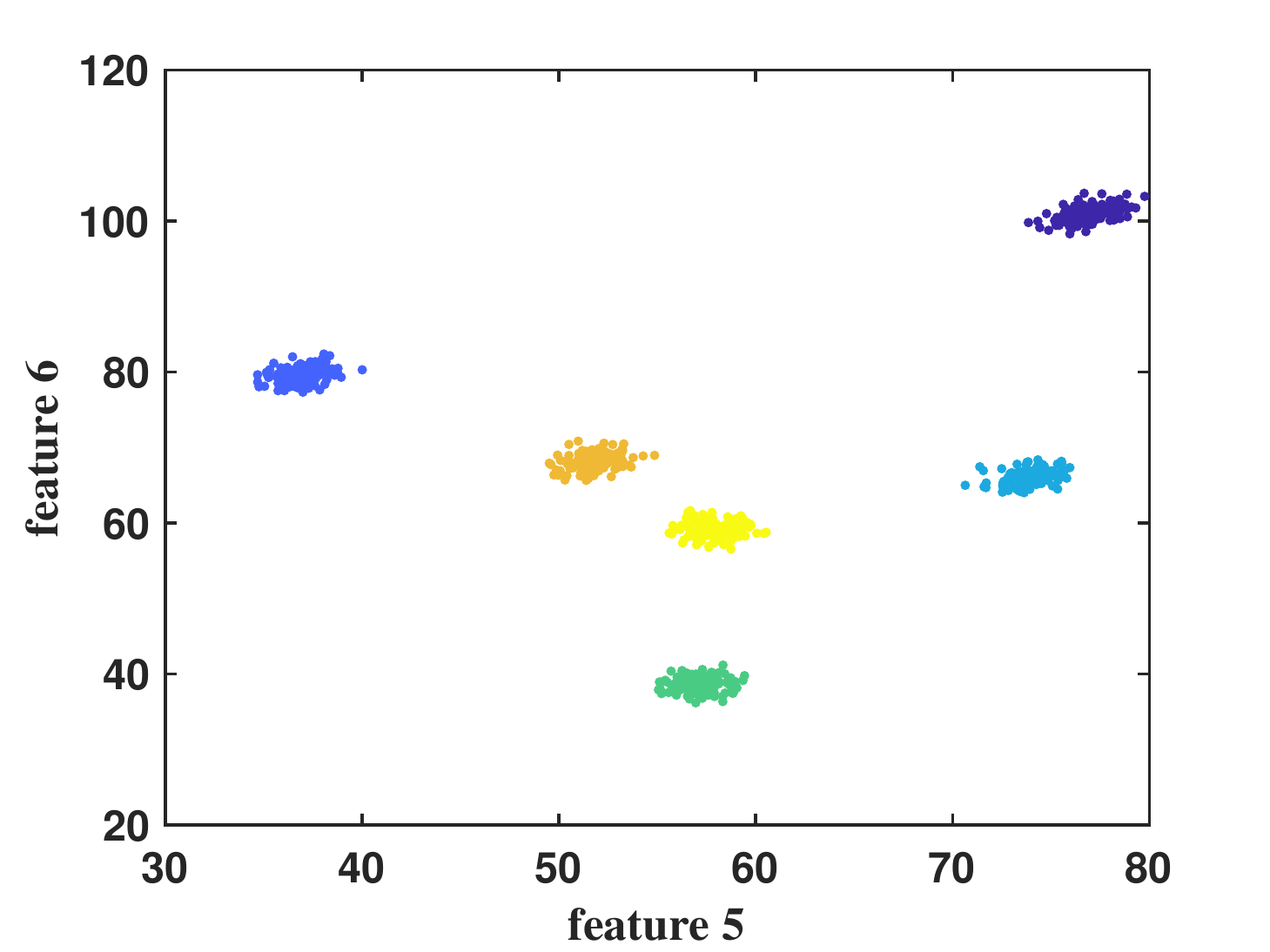}\\
\end{minipage}%
}%
\subfigure[\scriptsize Subspace4]{\label{fig:sy_s4}
\begin{minipage}[r]{0.11\textwidth}
\centering
  \includegraphics[width=\textwidth]{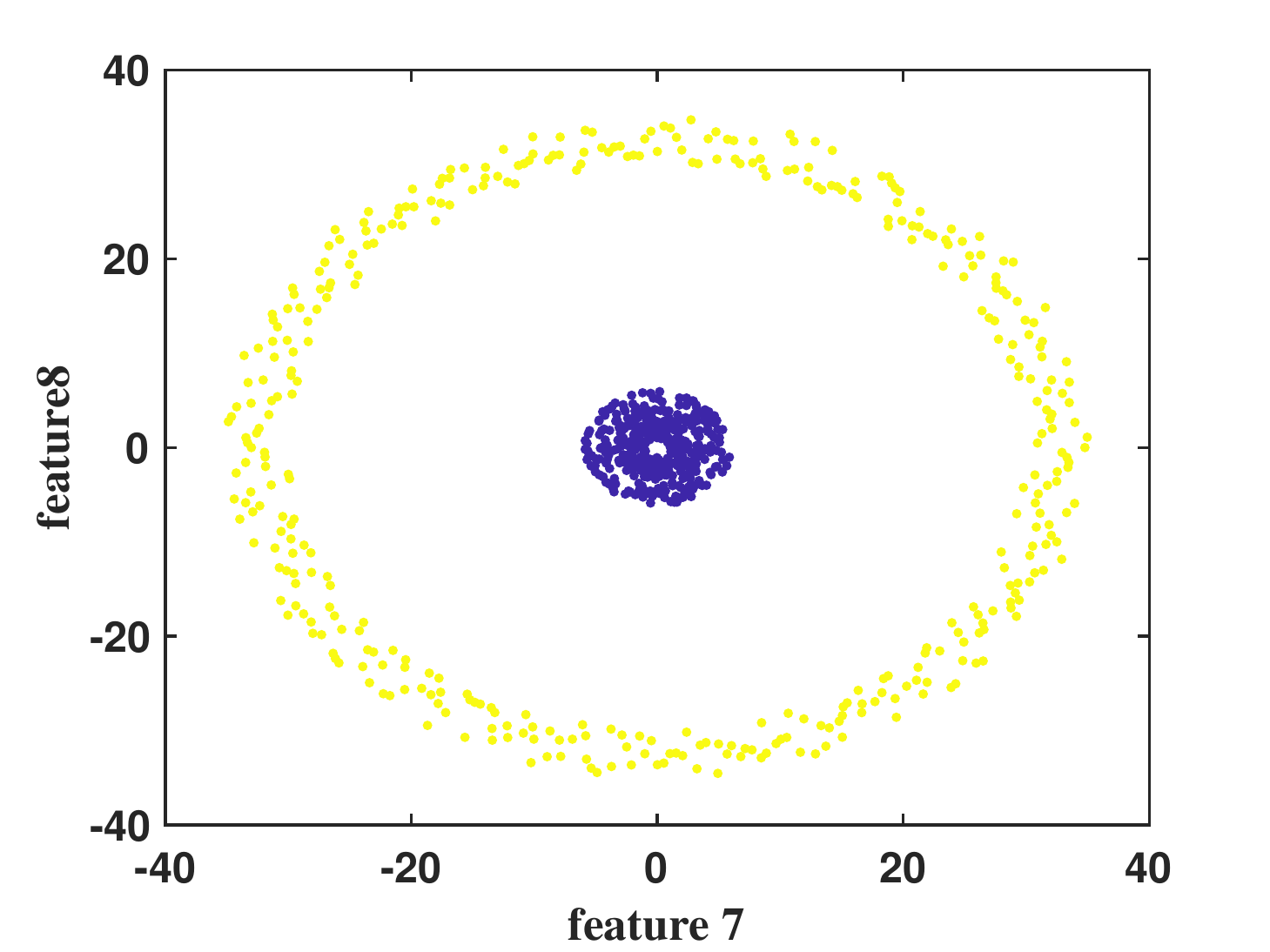}\\
\end{minipage}%
}%

\subfigure[\scriptsize View1]{\label{fig:sy_v1}
\begin{minipage}[r]{0.11\textwidth}
\centering
  \includegraphics[width=\textwidth]{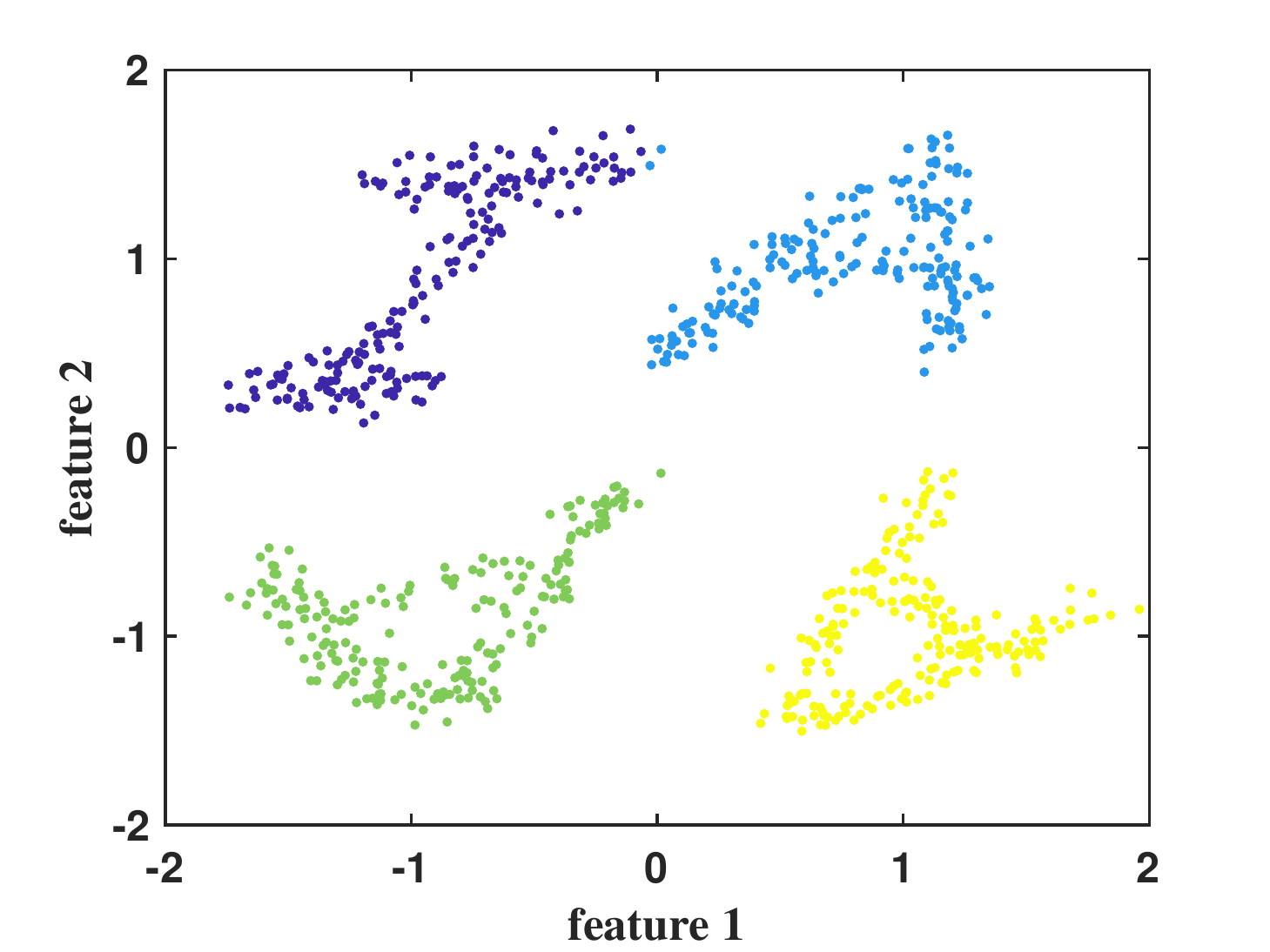}\\
\end{minipage}%
}
\subfigure[\scriptsize View2]{\label{fig:sy_v2}
\begin{minipage}[r]{0.11\textwidth}
\centering
  \includegraphics[width=\textwidth]{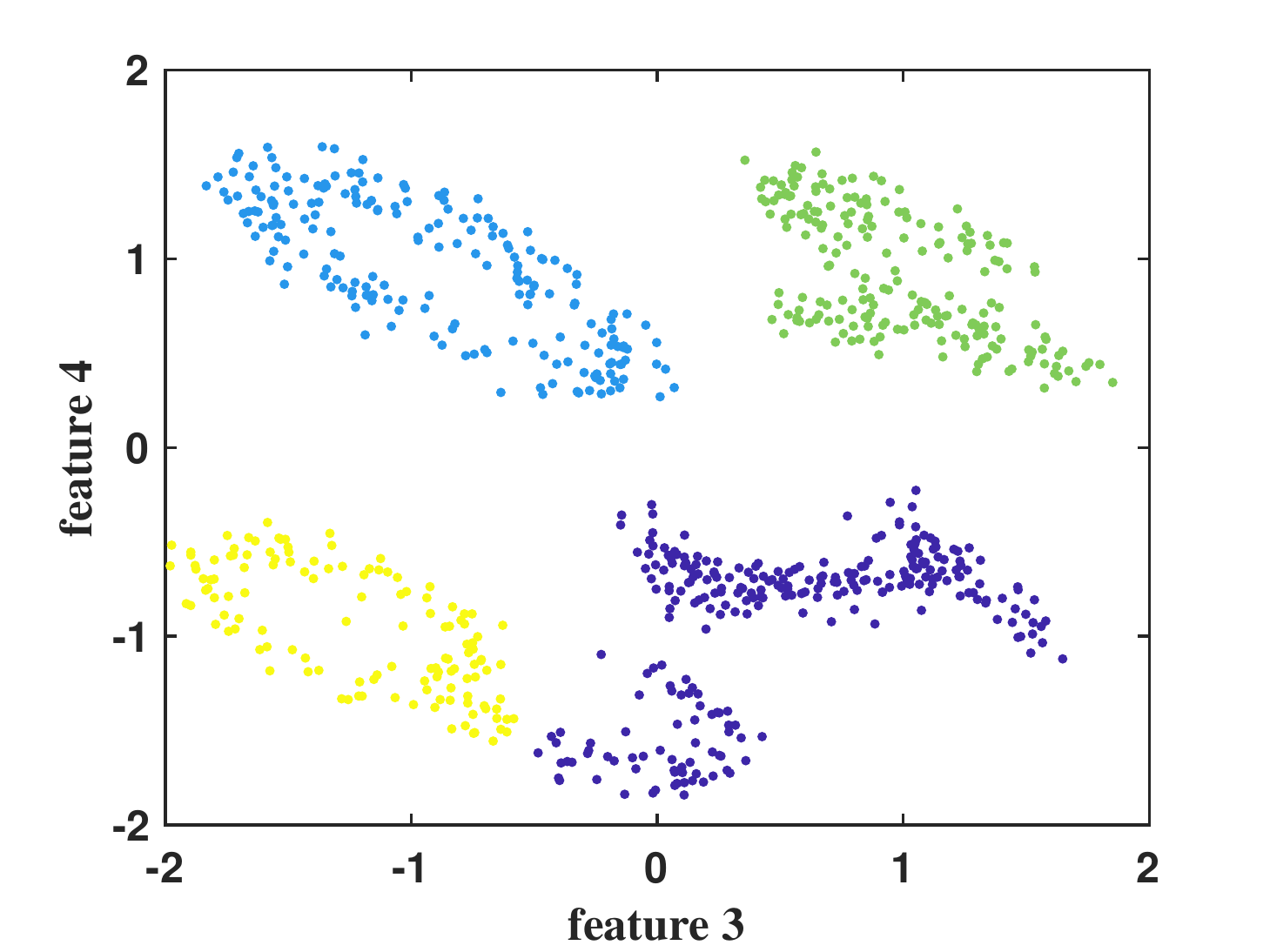}\\
\end{minipage}%
}%
\subfigure[\scriptsize View3]{\label{fig:sy_v3}
\begin{minipage}[r]{0.11\textwidth}
\centering
  \includegraphics[width=\textwidth]{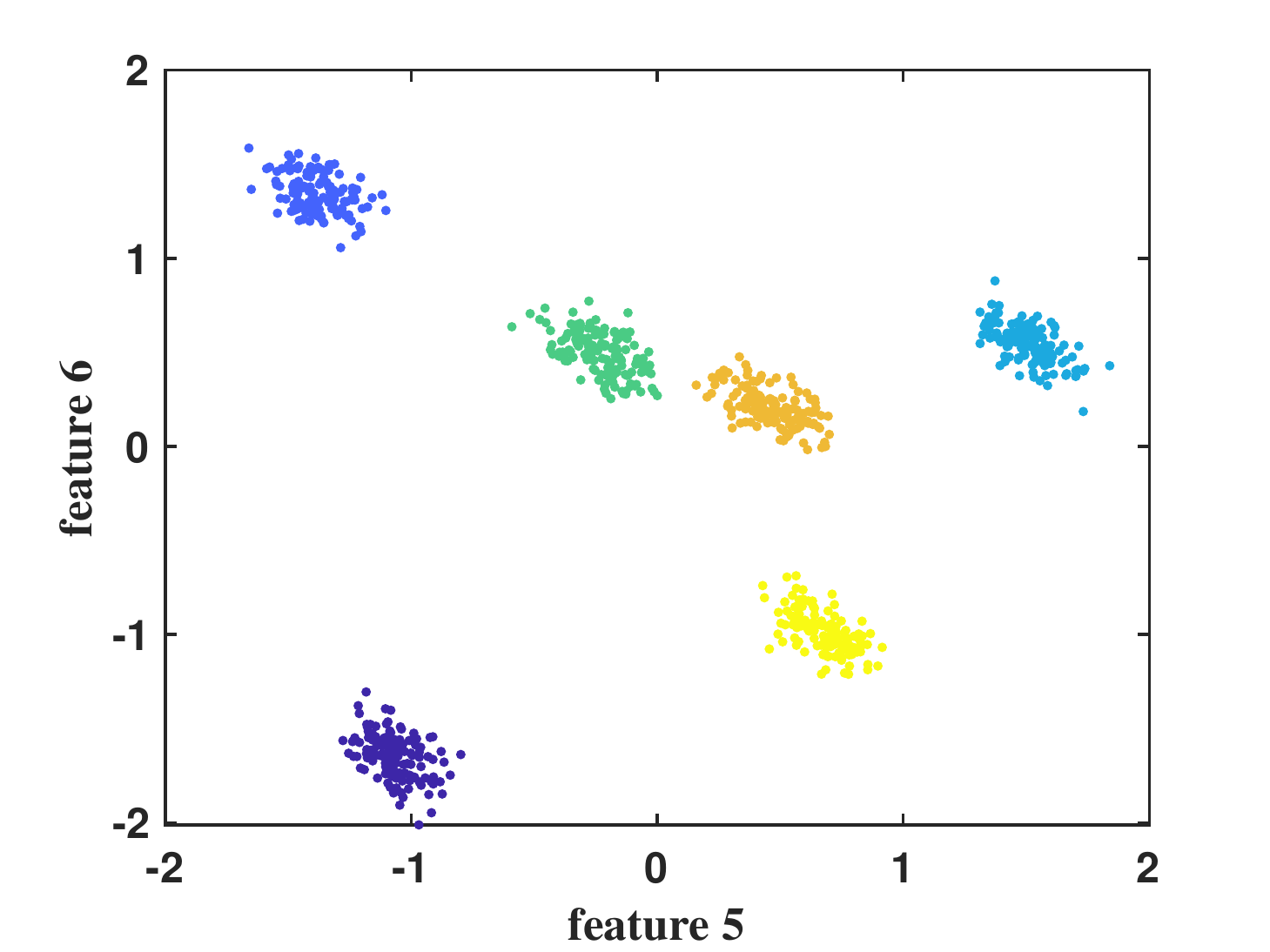}\\
\end{minipage}%
}%
\subfigure[\scriptsize View4]{\label{fig:sy_v4}
\begin{minipage}[r]{0.11\textwidth}
\centering
  \includegraphics[width=\textwidth]{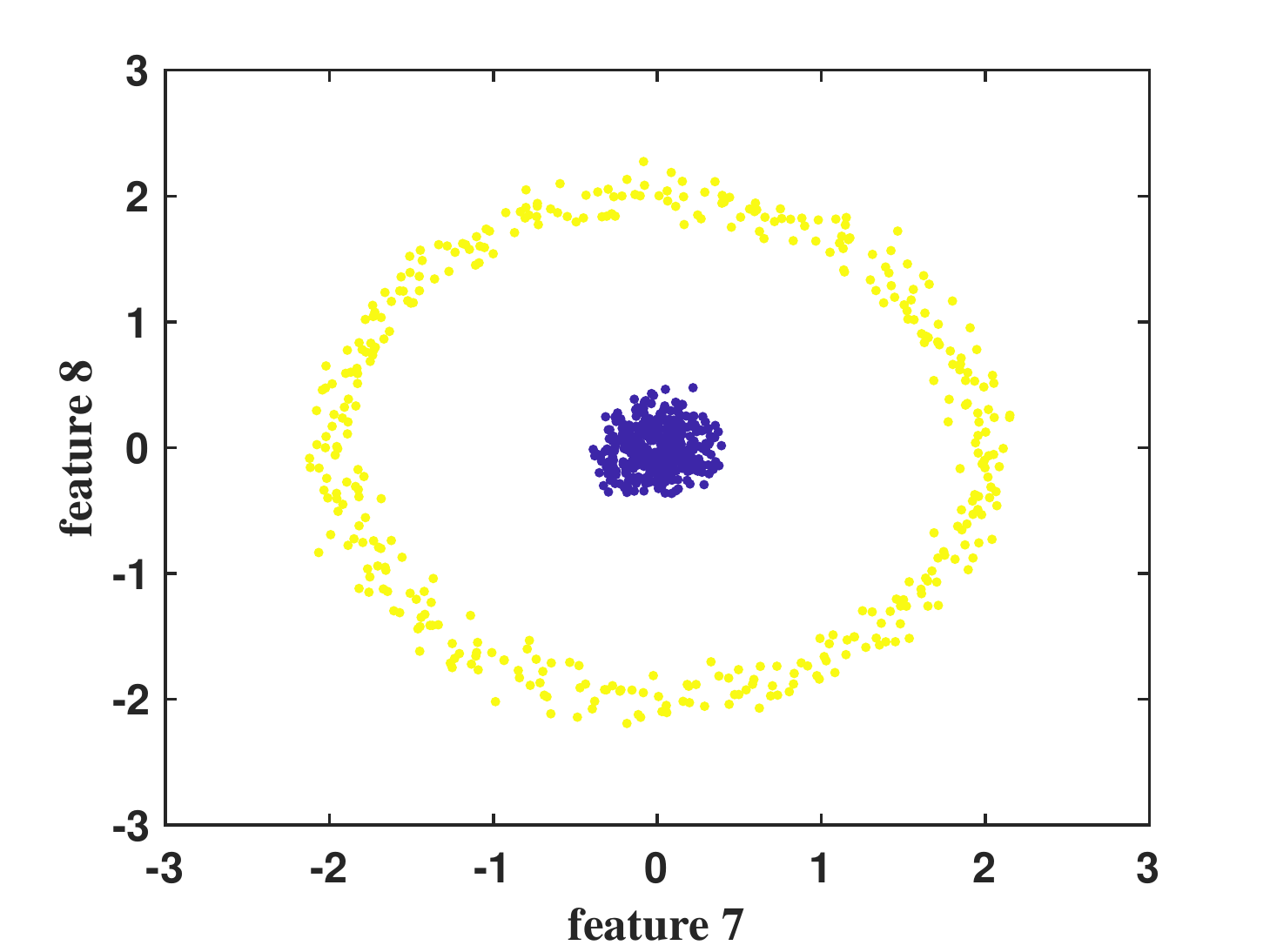}\\
\end{minipage}%
}%

\caption{Four different clusterings in four subspaces (a-d), and the four clusterings explored by MISC (e-h). } \label{fig:Sy_four}
\end{figure}

The first synthetic data contains four subspaces consisting of $800$ samples with $8$ features: the first subspace contains four clusters, corresponding to the  shapes of the digits `2',`0',`1',`9' (Fig. \ref{fig:sy_s1}); the second subspace also contains four clusters, corresponding to the shapes of the letters `A' (three shapes) and`I' (Fig. \ref{fig:sy_s2}); the third one contains  six clusters generated by a Gaussian distribution (Fig. \ref{fig:sy_s3}); the last one contains two clusters, which are non-linearly separable  (Fig. \ref{fig:sy_s4}). To ensure the non-redundancy among the four subspaces, we randomly permute the sample index in each subspace before merging them into a full space. Note that the synthetic data is diverse; it includes subspaces with the same scale, such as the first and the second subspaces, as well as  subspaces with different scales, such as the second, third, and fourth subspaces. We choose the Gaussian heat kernel as the kernel function and the kernel width is set to the standard variance $\sigma=sqrt(\sum_{i=1}^{n}\parallel \mathbf{X}_{\cdot i}-\overline{\mathbf{X}}\parallel^{2}/n)$. Following the set of GNMF in \cite{Cai2011Graph}, we use 0-1 weighting and adopt the neighborhood size $\epsilon=5$ to compute the graph adjacency matrix $\mathbf{P}$, and then set $\lambda=10$ in Eq. (\ref{Eq8}). We apply MISC on the first synthetic dataset and plot the found subspace views and clustering results in the last four subfigures of Fig. {\ref{fig:Sy_four}}.

The first view shown in Fig. {\ref{fig:sy_v1}} corresponds to the second original subspace; the second view shown in Fig. {\ref{fig:sy_v2}} corresponds to the first original subspace; the third view shown in Fig. {\ref{fig:sy_v3}} corresponds to the third original subspace; and the fourth view shown in Fig. {\ref{fig:sy_v4}} corresponds to the fourth original subspace. Due to the ISA procedure, the original feature space has been normalized and converted into the new space, so the four original subspaces are similar to the four subspaces found by MISC, but not identical. The relative position of each cluster in the new subspace is still the same as before, but the new subspaces are rotated and stretched because ICA tries to find subspaces which are  linear combinations of the original ones. For each subspace, we use KGSNMF to cluster the data. KGSNMF correctly identifies the clusters for the first, third, and fourth views; the second one is approximately close to the original one. Since KGSNMF accounts for the intrinsic geometric structure and for non-linearly separable clusters, it obtain good clustering results on both non-linearly separable  and spherical clusters.

\begin{figure}[h!tbp]
\centering
\scriptsize
\subfigure[\scriptsize $k$-means]{\label{fig:kmeans_Atom}
\begin{minipage}[r]{0.09\textwidth}
\centering
  \includegraphics[width=\textwidth]{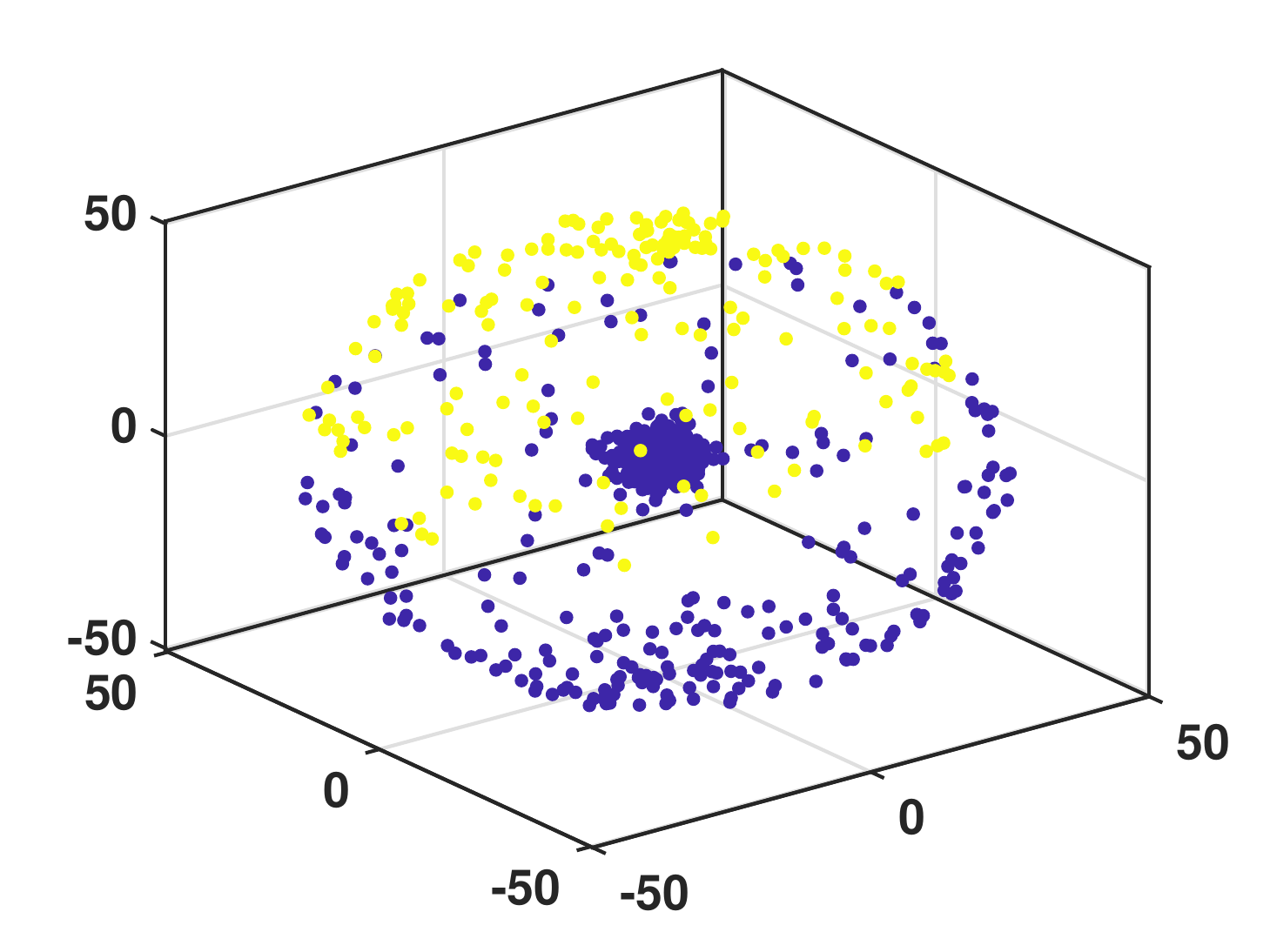}\\
\end{minipage}%
}%
\subfigure[\scriptsize SNMF]{\label{fig:semiCluster_Atom}
\begin{minipage}[r]{0.09\textwidth}
\centering
  \includegraphics[width=\textwidth]{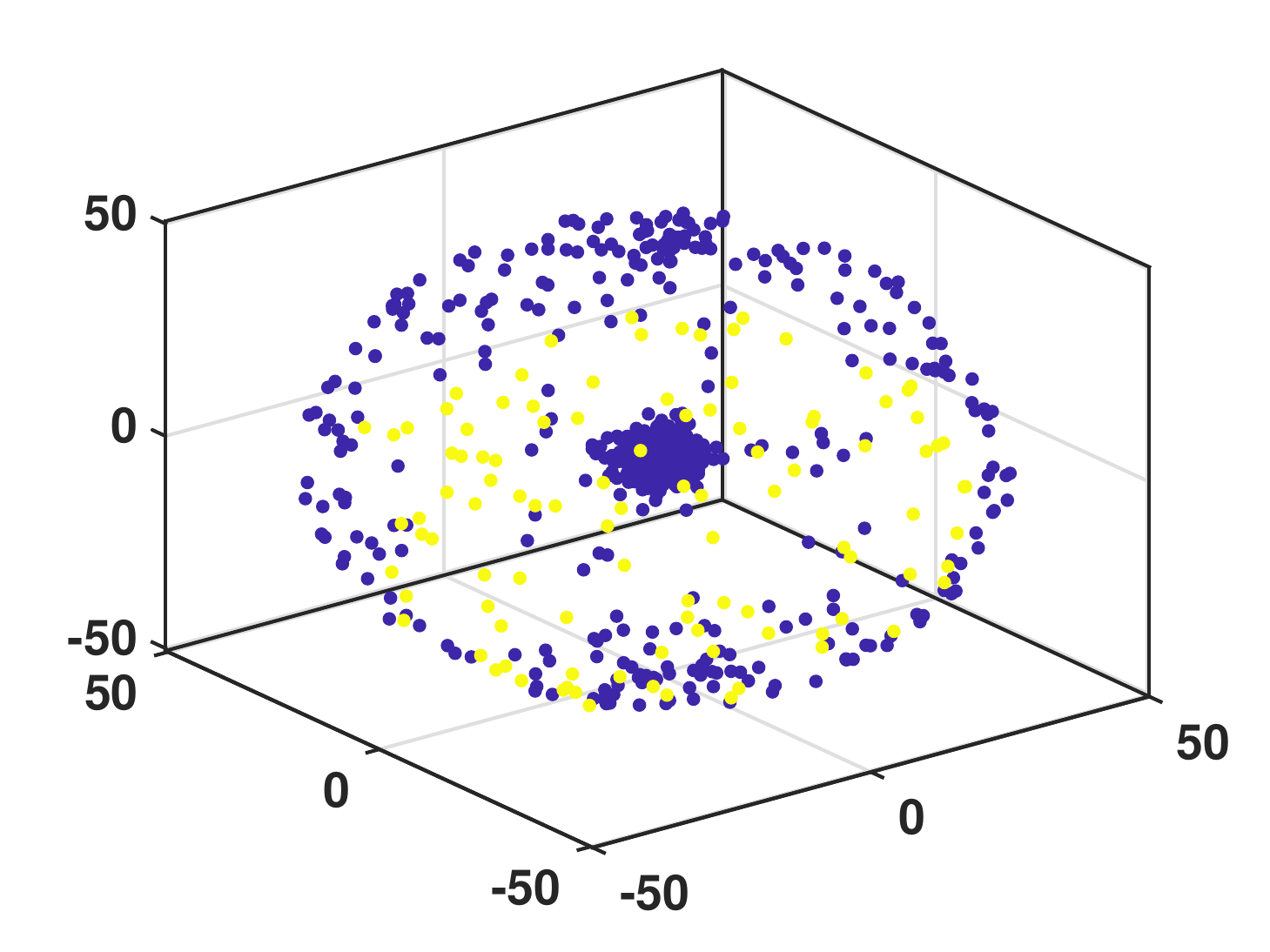}\\
\end{minipage}%
}%
\subfigure[\scriptsize GSNMF]{\label{fig:gnmfCluster_Atom}
\begin{minipage}[r]{0.09\textwidth}
\centering
  \includegraphics[width=\textwidth]{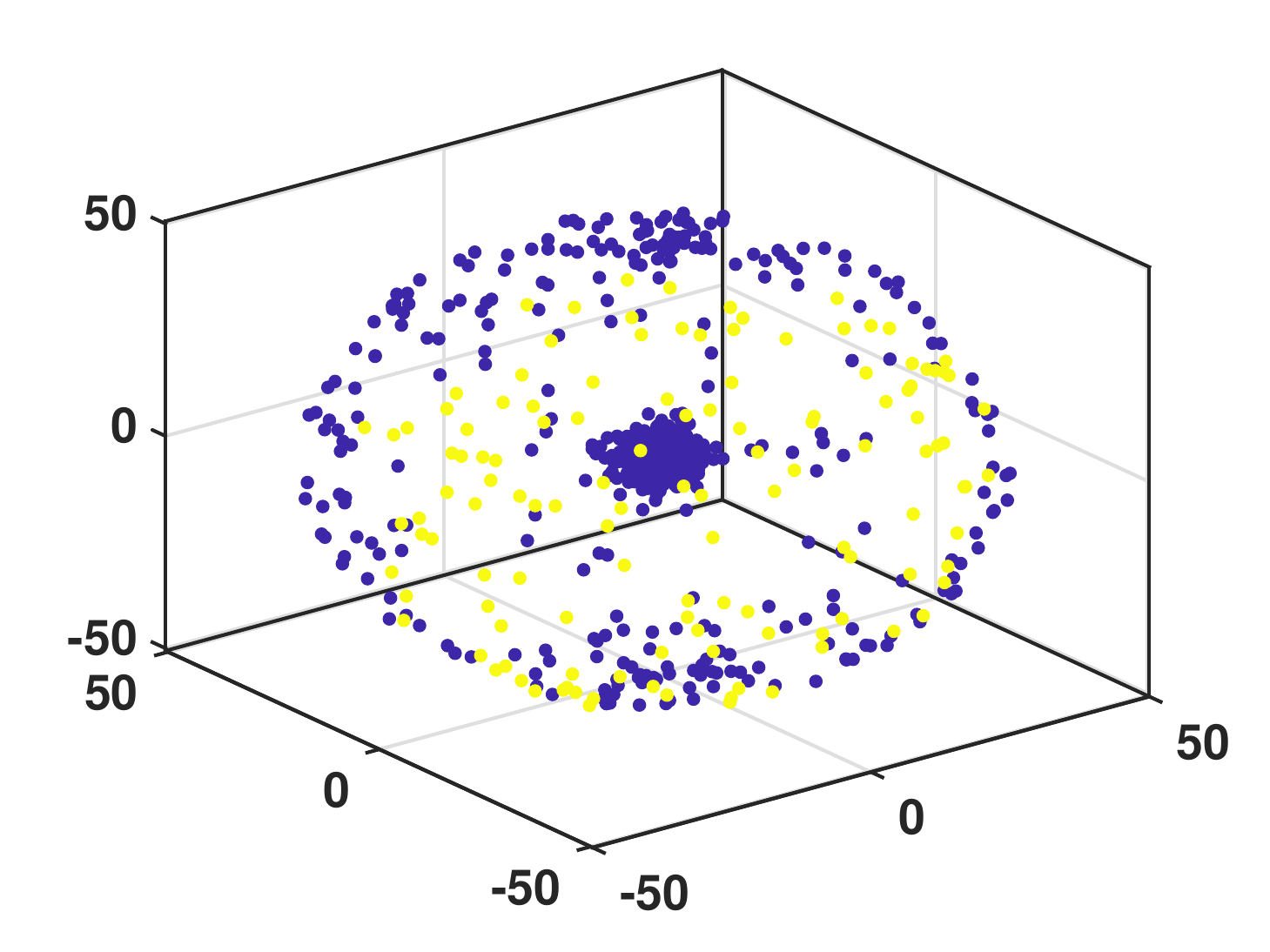}\\
\end{minipage}%
}
\subfigure[\scriptsize KSNMF]{\label{fig:kernelCluster_Atom}
\begin{minipage}[r]{0.09\textwidth}
\centering
  \includegraphics[width=\textwidth]{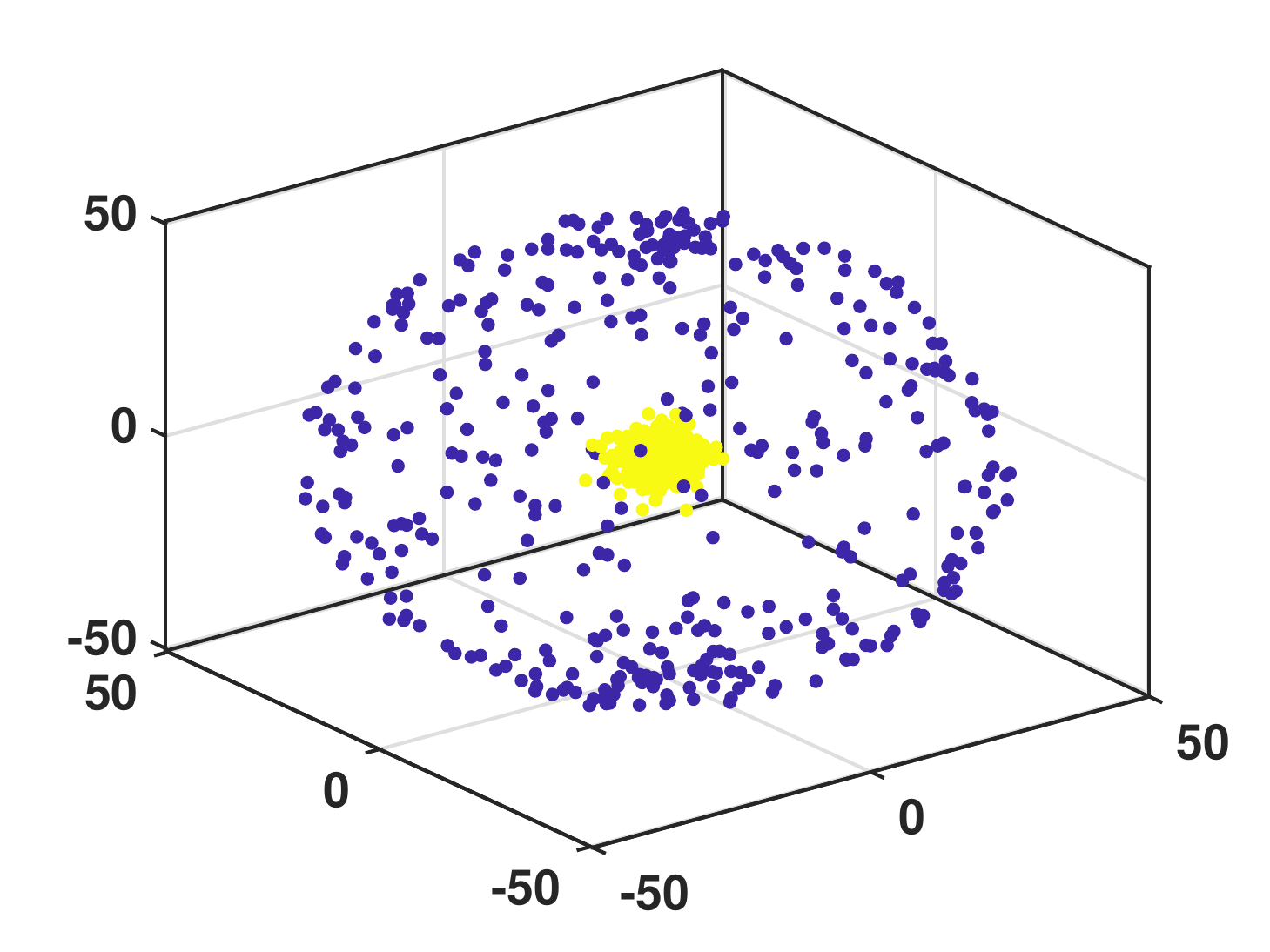}\\
\end{minipage}%
}%
\subfigure[\scriptsize KGSNMF]{\label{fig:kernelGnmfCluster_Atom}
\begin{minipage}[r]{0.09\textwidth}
\centering
  \includegraphics[width=\textwidth]{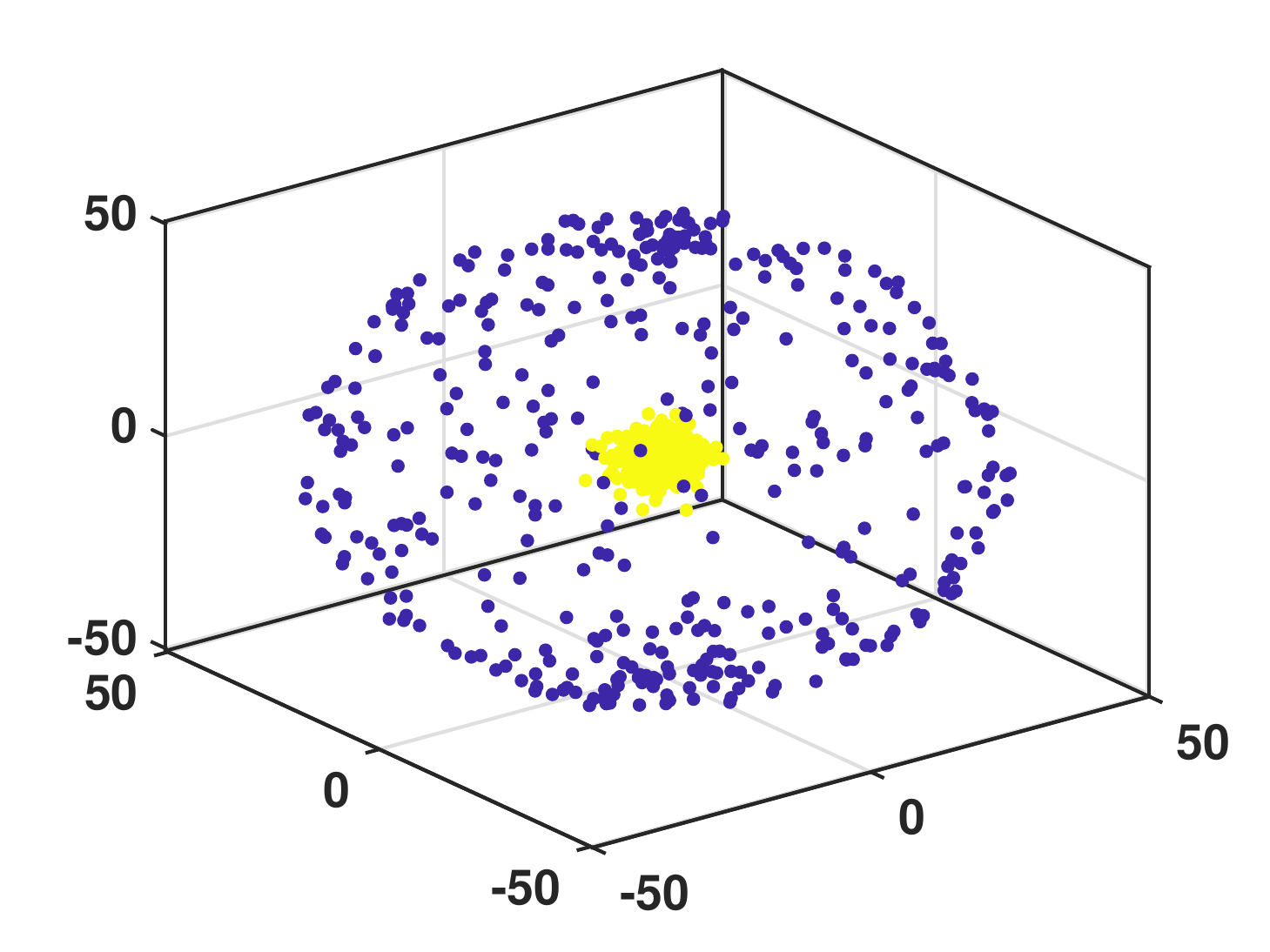}\\
\end{minipage}%
}%

\caption{Results of different clustering algorithms on the synthetic dataset Atom.} \label{fig:Sy_Atom}
\end{figure}

\begin{figure}[h!tbp]
\centering
\subfigure[\scriptsize $k$-means ]{\label{fig:kmeans_Lsun}
\begin{minipage}[r]{0.09\textwidth}
\centering
  \includegraphics[width=\textwidth]{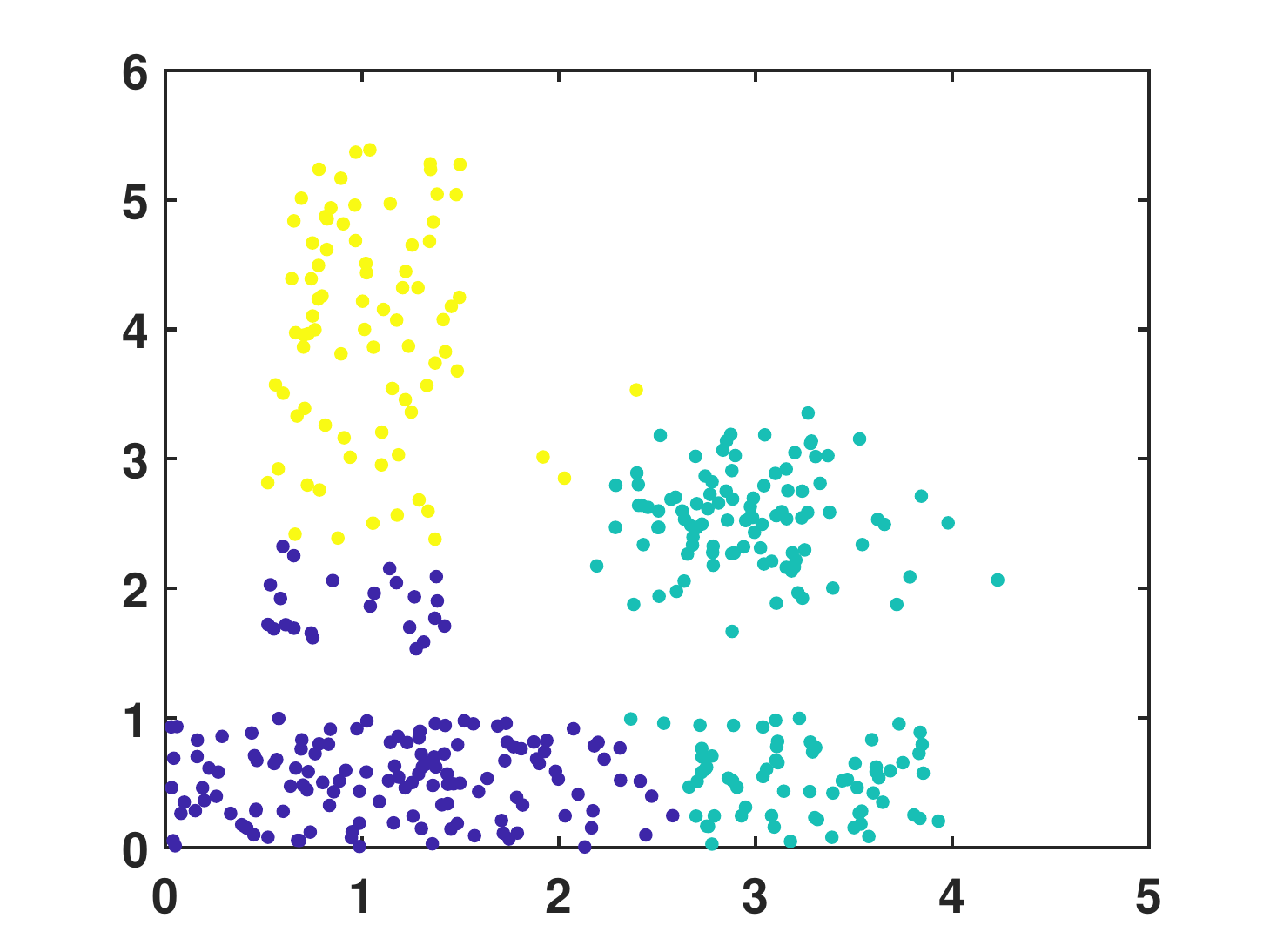}\\
\end{minipage}%
}%
\subfigure[\scriptsize SNMF]{\label{fig:semiCluster_Lsun}
\begin{minipage}[r]{0.09\textwidth}
\centering
  \includegraphics[width=\textwidth]{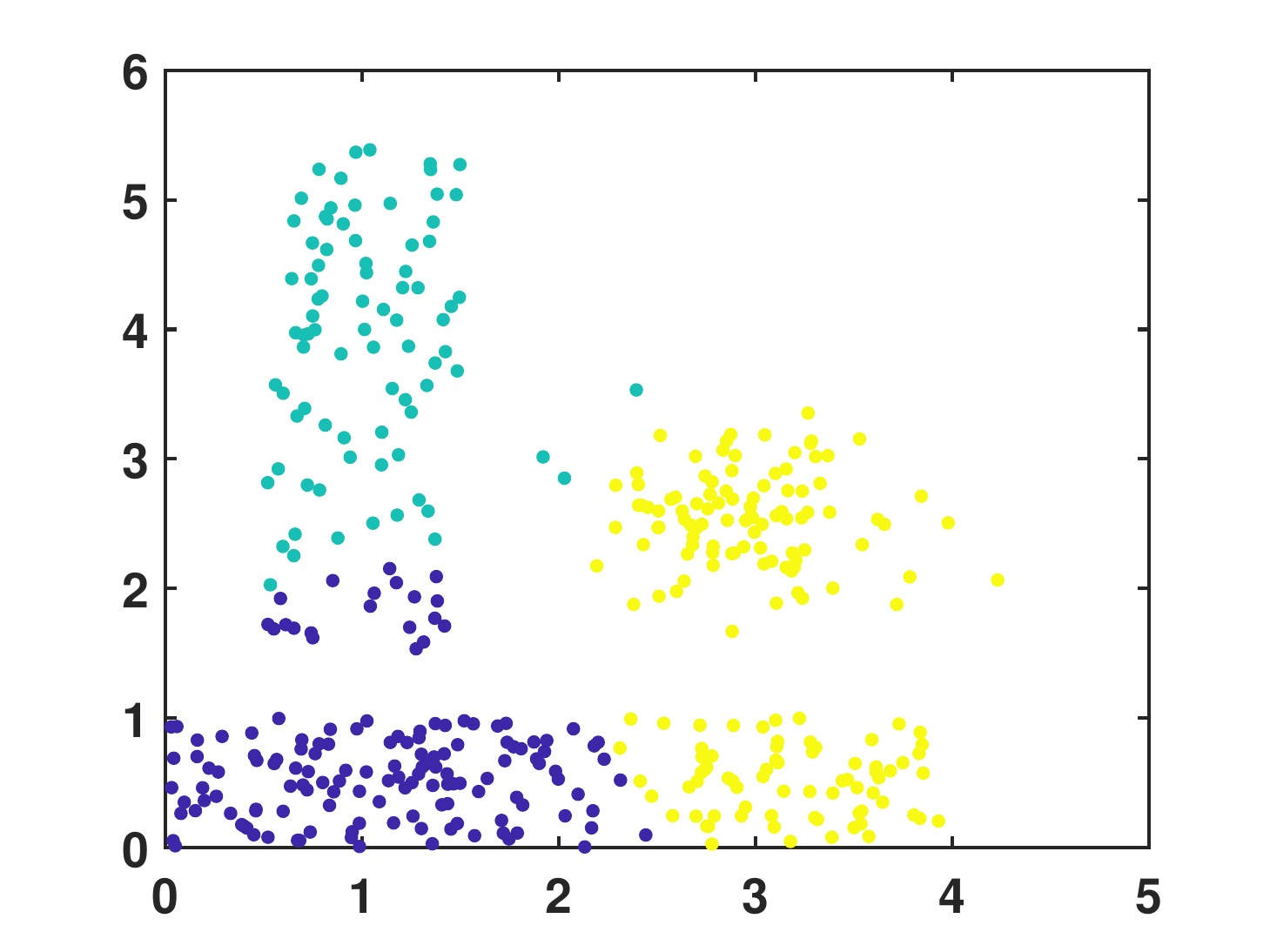}\\
\end{minipage}%
}%
\subfigure[\scriptsize GSNMF]{\label{fig:gnmfCluster_Lsun}
\begin{minipage}[r]{0.09\textwidth}
\centering
  \includegraphics[width=\textwidth]{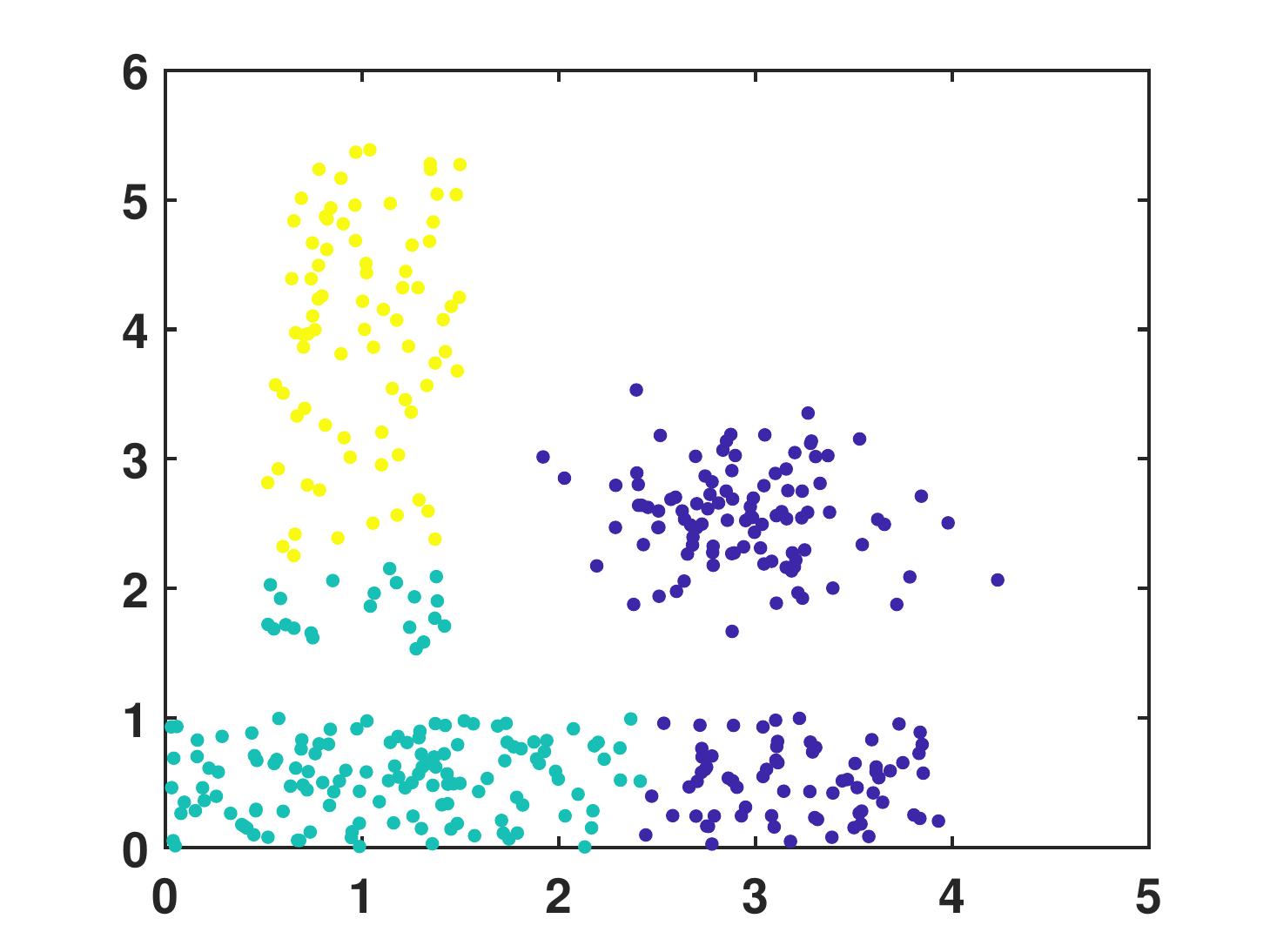}\\
\end{minipage}%
}
\subfigure[\scriptsize KSNMF]{\label{fig:kernelCluster_Lsun}
\begin{minipage}[r]{0.09\textwidth}
\centering
  \includegraphics[width=\textwidth]{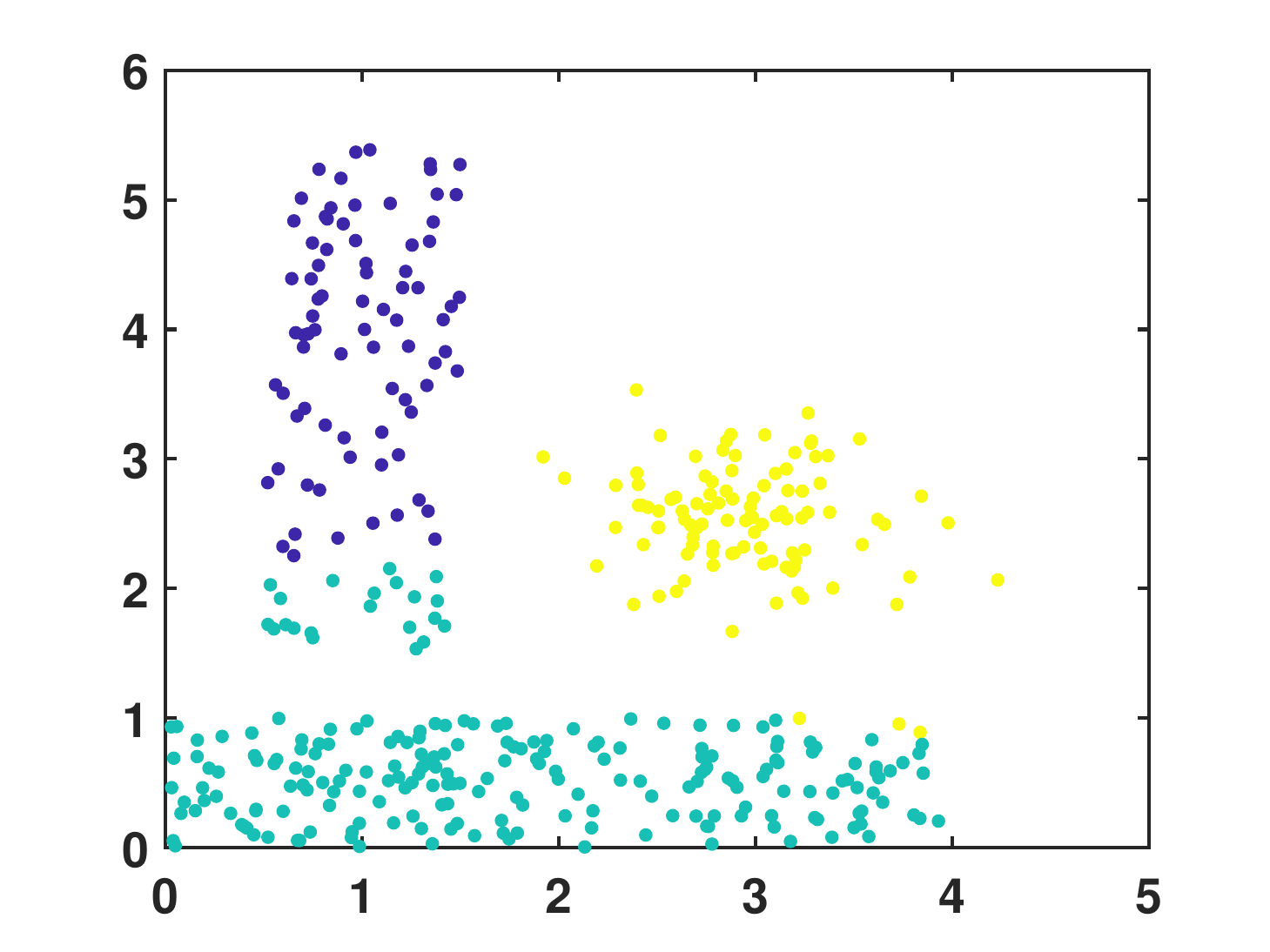}\\
\end{minipage}%
}%
\subfigure[\scriptsize KGSNMF]{\label{fig:kernelGnmfCluster_Lsun}
\begin{minipage}[r]{0.09\textwidth}
\centering
  \includegraphics[width=\textwidth]{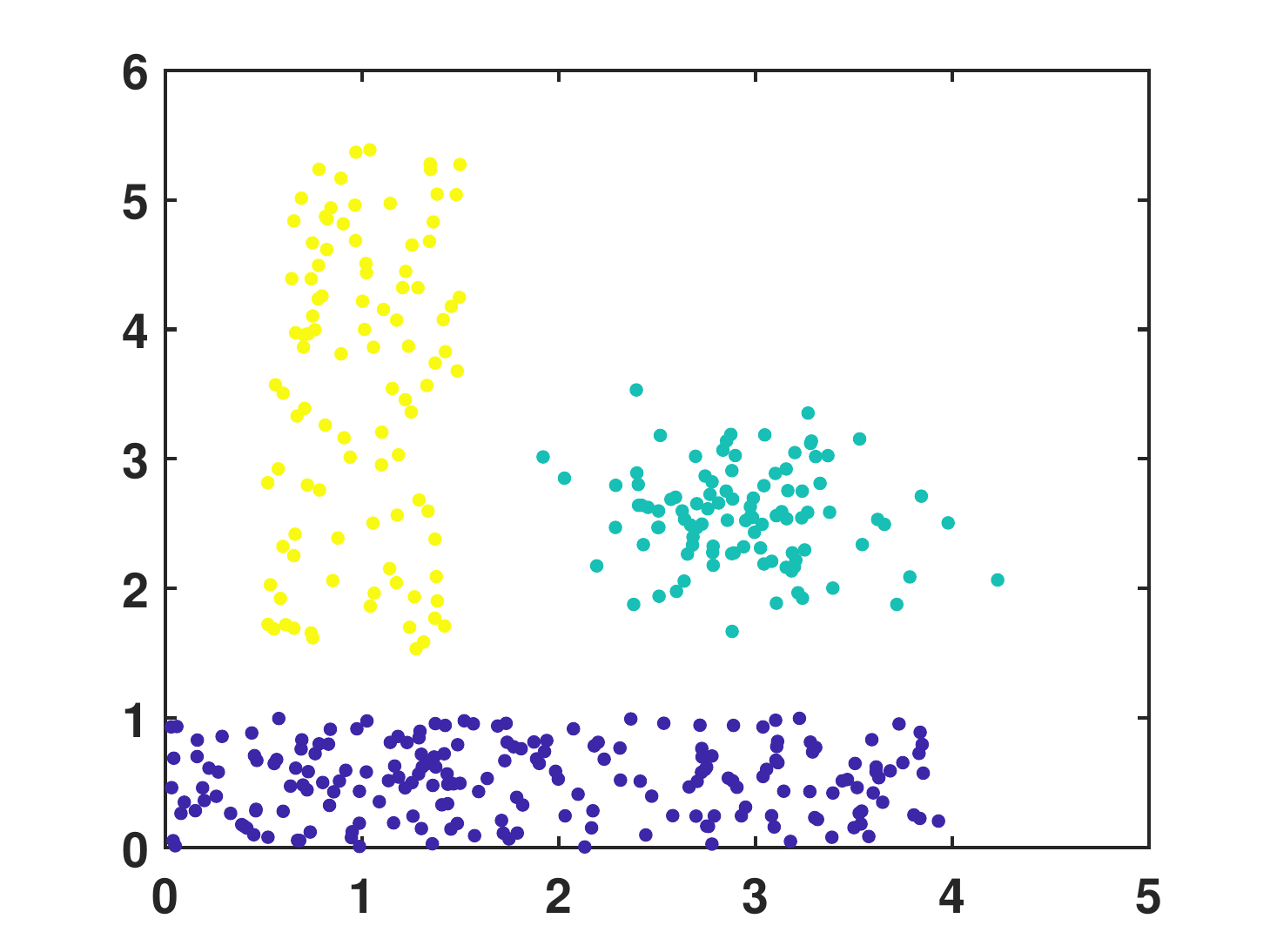}\\
\end{minipage}%
}%

\caption{Results of different clustering algorithms on the synthetic dataset Lsun.} \label{fig:Sy_Lsun}
\end{figure}

The second and third synthetic datasets are collected from the Fundamental Clustering Problem Suite (FCPS)\footnote{http://www.uni-marburg.de/fb12/datenbionik/downloads/FCPS}. We  use them to investigate whether  KGSNMF achieves a better clustering performance than SNMF. Atom, the second synthetic dataset, consists of $800$ samples with three features. It contains two nonlinearly separable clusters with different variance  as shown in Fig. {\ref{fig:Sy_Atom}}. Lsun, the third synthetic dataset, consists of $400$ samples with two features. It contains three clusters with different variance and inner-cluster distance as shown in Fig. {\ref{fig:Sy_Lsun}}. We  choose a Gaussian kernel and set $\lambda=10$ for KGSNMF and GSNMF as before. The clustering results on Atom are plotted in Fig. {\ref{fig:Sy_Atom}}, and we can see that both KSNMF and KGSNMF correctly separate the two clusters, while  $k$-means, SNMF, and GSNMF do not. This is because the introduced kernel function could map the nonlinearly  separable space to a high-dimensional linearly separable space. The clustering results for Lsun are shown in Fig. {\ref{fig:Sy_Lsun}}. $k$-means, SNMF, GSNMF, and KSNMF do not cluster the data very well. $k$-means, SNMF, and GSNMF are all influenced by the distribution of the clusters at the bottom. KSNMF can mitigate the impact, but it still cannot perfectly separate the clusters, whereas KGSNMF can do the job correctly. Overall,  KSNMF achieves good clustering results especially on nonlinearly separable clusters, such as on Atom. The impact of different structures could be alleviated to some extent on both linearly and nonlinearly separable data. The embedded graph regularized term can better represent the details of the intrinsic geometry of the data; as such KGSNMF obtains  better clustering results than KSNMF.

\subsection{Experiments on real-world datasets}
\begin{figure}[h!tbp]
  \centering
  \includegraphics[width=0.2\textwidth]{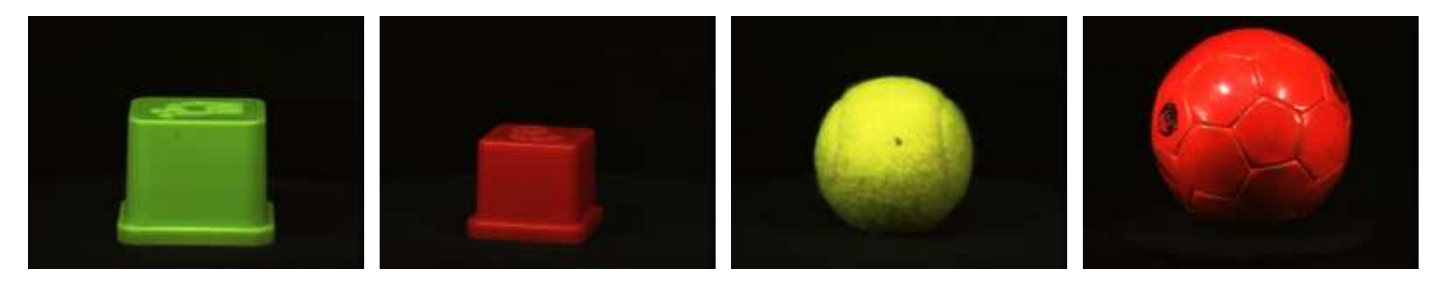}\\
  \caption{Four objects of different shapes (box and ball) and colors (green and red) from ALOI.}\label{fig:org_aloi}
\end{figure}

\begin{figure}[htbp]
  \centering
  \includegraphics[width=0.3\textwidth]{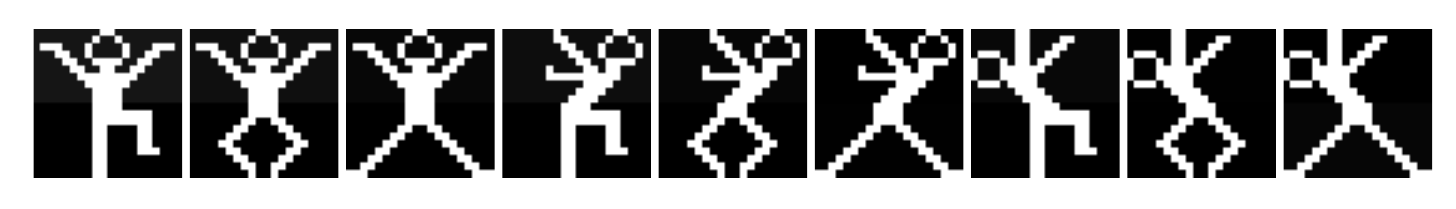}\\
  \caption{Nine raw samples of the Dancing Stick Figures.}\label{fig:org_dsf}
\end{figure}

We test MISC on four real-world datasets wildly used for multiple clustering, including a color image dataset, two gray image datasets, and a text dataset.
\begin{itemize}
  \item \textbf{Amsterdam Library of Object Images dataset.} The ALOI dataset\footnote{http://aloi.science.uva.nl/} consists of images of 1000 common objects taken from different angles and under various illumination conditions. We have chosen four objects: green box, red box, tennis ball, and red ball, with different colors and shapes from different viewing directions for a total of 288 images (Fig. {\ref{fig:org_aloi}}). Following the preprocessing in \cite{Dalal2005Histograms}, we extracted 840 features\footnote{https://github.com/adikhosla/feature-extraction} and further applied Principle Component Analysis (PCA) to reduce the number of features to 49, which retain more than 90\% variance of the original data.
  \item \textbf{Dancing Stick Figures dataset}. The DSF dataset \cite{G2014SMVC} consists of 900 samples of $20\times20$ images with random noise across nine stick figures. (Fig. \ref{fig:org_dsf}). The nine raw stick figures are obtained by arranging in three different positions the upper and lower body; this provides two views for the dataset. As for the ALOI, we also applied PCA, and retained more than 90\% of the data's variance as preprocessing.
  \item \textbf{CMUface dataset.} The CMUface dataset\footnote{http://archive.ics.uci.edu/ml/datasets.html} contains 640 grey $32\times20$ images of 20 individuals with varying poses (up, straight, right, and left). As such, it can be clustered either by identity or by pose. Again, we apply PCA to reduce the dimensionality while retaining more than 90\% of the data's variance.
  \item \textbf{WebKB dataset} The WebKB dataset\footnote{http://www.cs.cmu.edu/~webkb/} contains html documents from four universities: Cornell University; University of Texas, Austin; University of Washington; and University of Wisconsin, Madison. The pages are additionally labeled as being from 4 categories: course, faculty, project, and student. We preprocessed the data by removing rare words, stop words, and words with a small variance, retaining 1041 samples and 456 words.
\end{itemize}

We compare MISC with MetaC, MSC, OSC, COALA, De-$k$means, ADFT, MNMF, mSC, NBMC, and NBMC-OFV (all methods are discussed in the related work section). The input parameters of these algorithms were set or optimized as suggested by the authors. We also set the number of subspaces as 2 and the number of clusters as that of true labels of CMUface and WebKB datasets, respectively.

\begin{figure}[h!tbp]
\centering
\subfigure[Subspace1: shape]{\label{fig:aloi_s1}
\begin{minipage}[r]{0.2\textwidth}
\centering
  \includegraphics[width=0.7\textwidth]{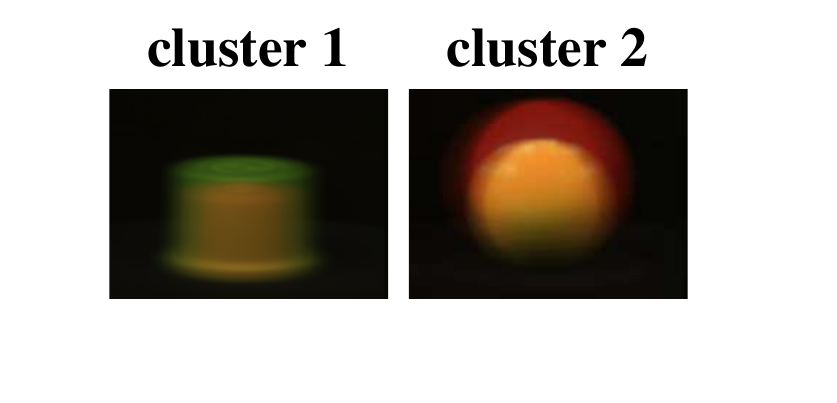}\\
\end{minipage}%
}
\subfigure[Subspace2: color]{\label{fig:aloi_s2}
\begin{minipage}[r]{0.2\textwidth}
\centering
  \includegraphics[width=0.7\textwidth]{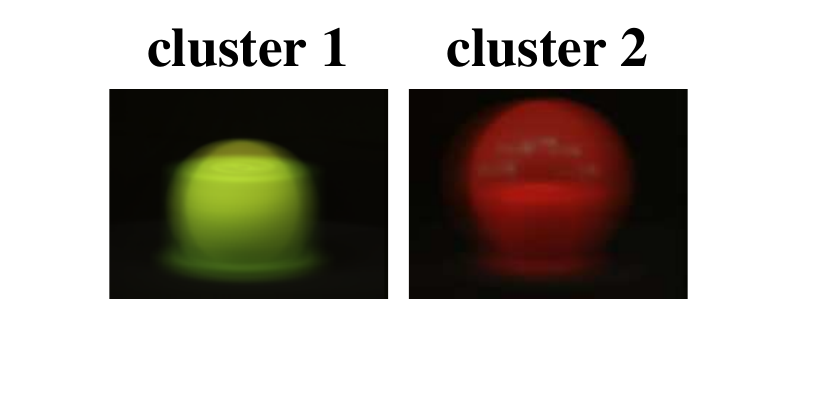}\\
\end{minipage}%
}%
\caption{ALOI dataset: Mean images of the clusters in two subspaces detected by MISC from the perspective of shape (a) and color (b).} \label{fig:aloi_MISC}
\end{figure}

\begin{figure}[h!tbp]
\centering
\subfigure[Subspace1: upper-body]{\label{fig:dsf_s1}
\begin{minipage}[r]{0.2\textwidth}
\centering
  \includegraphics[width=0.8\textwidth]{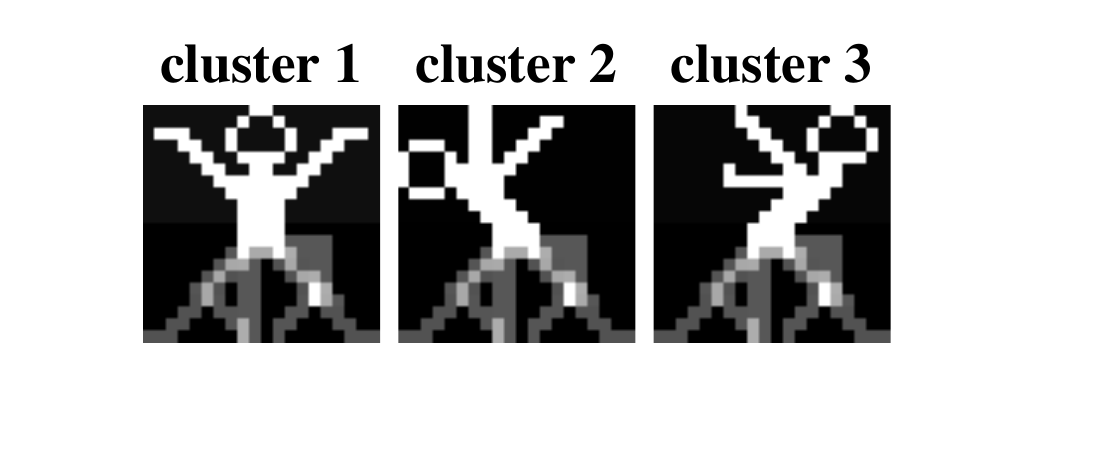}\\
\end{minipage}%
}
\subfigure[Subspace2: lower-body]{\label{fig:dsf_s2}
\begin{minipage}[r]{0.2\textwidth}
\centering
  \includegraphics[width=0.8\textwidth]{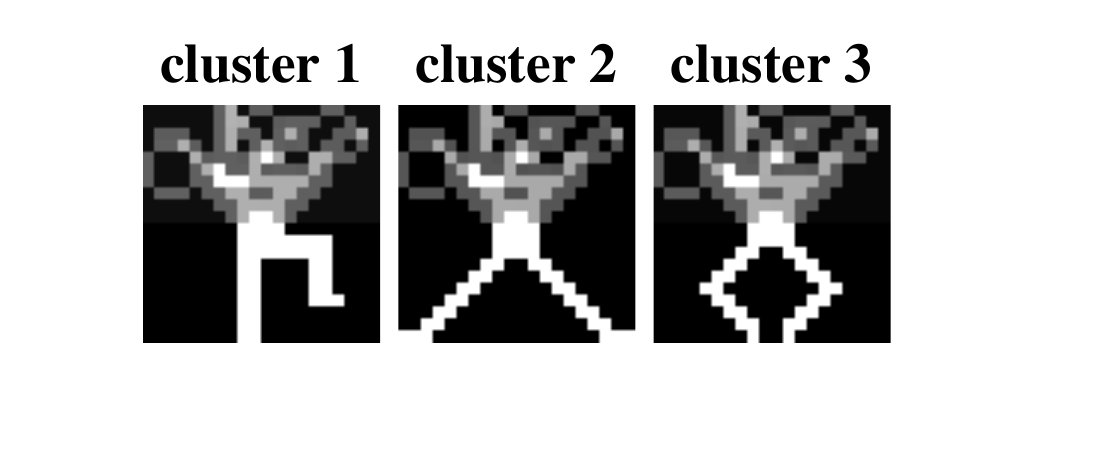}\\
\end{minipage}%
}%
\caption{DSF dataset: Mean images of the clusters in two subspaces detected by MISC from the perspective of the upper-body (a) and lower-body (b).} \label{fig:dsf_MISC}
\end{figure}

\begin{figure}[h!t!bp]
\centering
\subfigure[Subspace1: identity]{\label{fig:Face_ID}
\begin{minipage}[r]{0.2\textwidth}
\centering
  \includegraphics[width=0.8\textwidth,]{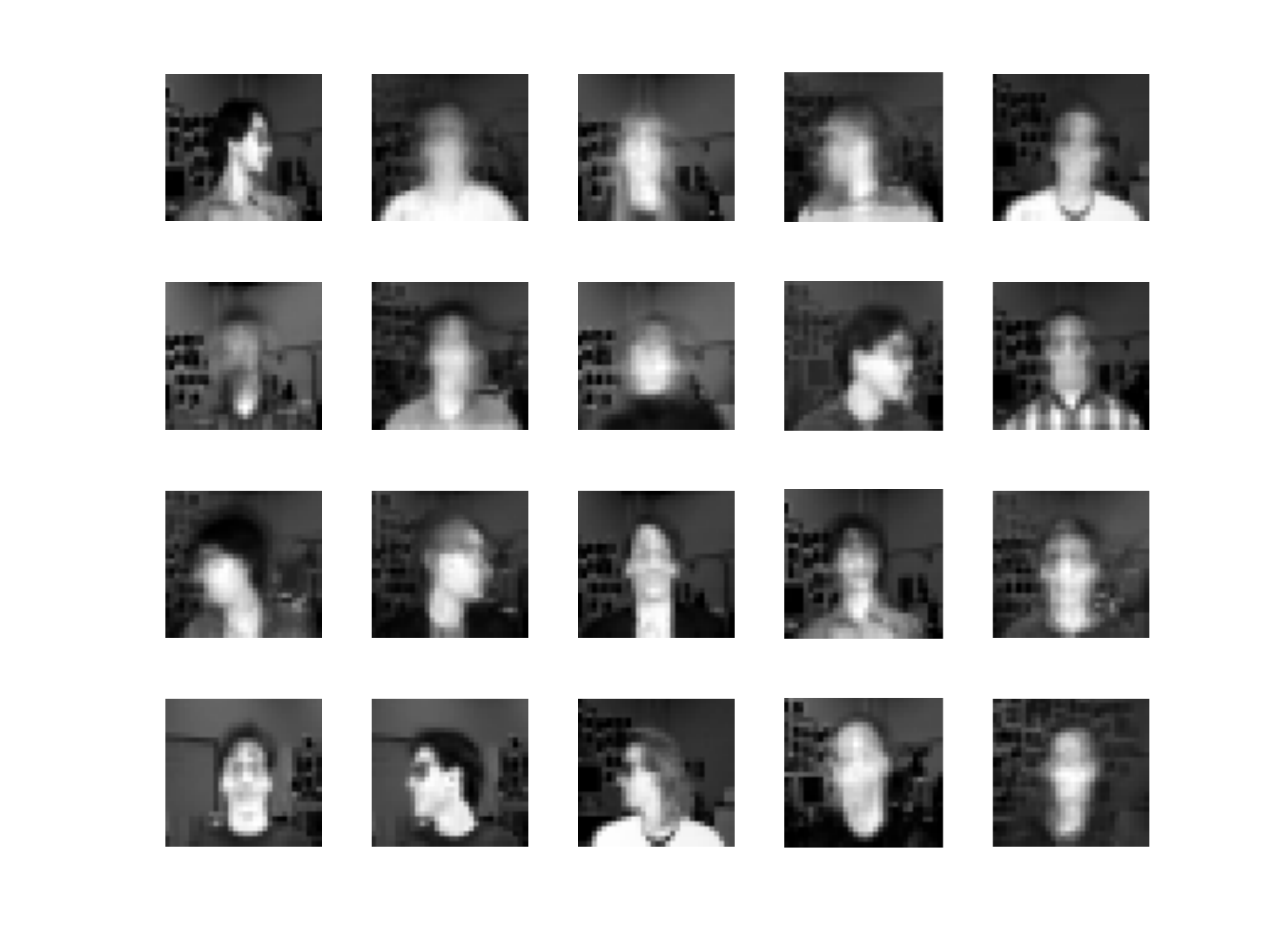}\\
\end{minipage}%
}
\subfigure[Subspace2: pose]{\label{fig:Face_Pose}
\begin{minipage}[r]{0.2\textwidth}
\centering
  \includegraphics[width=0.8\textwidth]{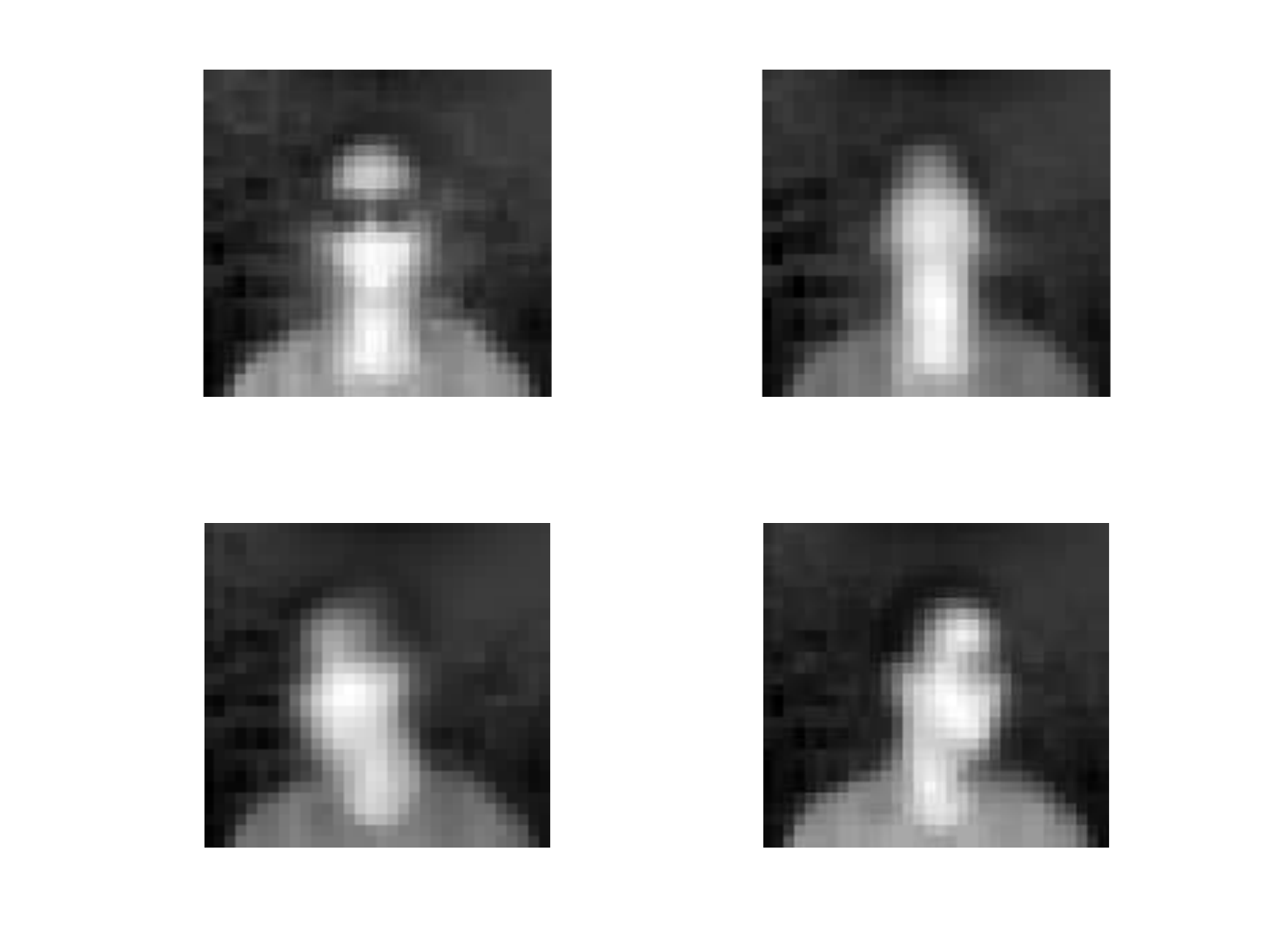}\\
\end{minipage}%
}%
\caption{The mean image of the clusters of two clusterings in two subspaces of CMUface detected by MISC from the perspective of identity (a) and pose (b).}\label{fig:Face_MISC}
\end{figure}

We visualize the clustering results of MISC for the first three image datasets in Figs. \ref{fig:aloi_MISC}-\ref{fig:Face_MISC}, and  use the widely-known F1-measure (F1) and normalized mutual information (NMI) to evaluate the quality of the clusterings. Since we don't know which view the clustering corresponds to, we compare each clustering  with the true label under each view, and finally compute the confusion matrix and report the results (average of ten independent repetitions) in Table \ref{tab:F1}.


\begin{table*}[h!t]
\renewcommand\arraystretch{0.72} 
\scriptsize
\caption{F1 and NMI confusion matrix  (Mean$\pm$Std). $\mathcal{C}_1$ and $\mathcal{C}_2$ indicate two clusterings of the same data. $\bullet / \circ$ indicates whether MISC is statistically (according to pairwise $t$-test at 95\% significance level) superior/inferior to the other method. The bold numbers represent the best results.}
\centering
\begin{tabular}{l l| l l| l l| l l| l l}
\hline
\multicolumn{2}{c|}{\multirow{2}[3]{*}{\textbf{F1}}}
& \multicolumn{2}{c|}{ALOI} & \multicolumn{2}{c|}{DSF} & \multicolumn{2}{c|}{CMUface} & \multicolumn{2}{c}{WebKB} \\
\cline{3-10}
\multicolumn{2}{c|}{} &\multicolumn{1}{c}{Shape} &\multicolumn{1}{c|}{Color} &\multicolumn{1}{c}{Upper-body} &\multicolumn{1}{c|}{Lower-body} &\multicolumn{1}{c}{Identity} &\multicolumn{1}{c|}{Pose} &\multicolumn{1}{c}{University} &\multicolumn{1}{c}{Category} \\
\hline
\multirow{2}{*}{MetaC}
    & $\mathcal{C}_1$    & 0.783$\pm$0.022$\bullet$ & 0.636$\pm$0.022$\bullet$ & 0.871$\pm$0.019$\bullet$ & 0.433$\pm$0.019$\bullet$ & 0.234$\pm$0.019$\bullet$	& 0.284$\pm$0.025$\bullet$	& 0.473$\pm$0.021$\bullet$	& 0.442$\pm$0.014$\bullet$	\\
    & $\mathcal{C}_2$    & 0.716$\pm$0.020$\bullet$ & 0.616$\pm$0.018$\bullet$ & 0.610$\pm$0.025$\bullet$ & 0.622$\pm$0.024$\bullet$ & 0.542$\pm$0.025$\bullet$	& 0.130$\pm$0.024$\bullet$	& 0.402$\pm$0.028$\bullet$	& 0.474$\pm$0.018$\bullet$	\\
\hline                                                                                                                               				
\multirow{2}{*}{MSC}                                                                                                                 				
    & $\mathcal{C}_1$    & 0.759$\pm$0.021$\bullet$ & 0.605$\pm$0.014$\bullet$ & 0.738$\pm$0.018$\bullet$ & 0.476$\pm$0.020$\bullet$ & 0.592$\pm$0.012$\bullet$	& 0.115$\pm$0.030$\bullet$	& 0.463$\pm$0.018$\bullet$	& 0.502$\pm$0.026	\\
    & $\mathcal{C}_2$    & 0.597$\pm$0.019$\bullet$ & 0.799$\pm$0.017$\bullet$ & 0.498$\pm$0.023$\bullet$ & 0.681$\pm$0.019$\bullet$ & 0.23$\pm$0.0180$\bullet$	& 0.386$\pm$0.017$\bullet$	& 0.456$\pm$0.019$\bullet$	& 0.513$\pm$0.018$\circ$	\\
\hline                                                                                                                               				
\multirow{2}{*}{OSC}                                                                                                                 				
    & $\mathcal{C}_1$    & 0.681$\pm$0.020$\bullet$ & 0.732$\pm$0.018$\bullet$ & 0.683$\pm$0.023$\bullet$ & 0.482$\pm$0.021$\bullet$ & 0.343$\pm$0.013$\bullet$	& 0.292$\pm$0.015$\bullet$	& 0.462$\pm$0.020$\bullet$	& 0.490$\pm$0.020$\bullet$	\\
    & $\mathcal{C}_2$    & 0.732$\pm$0.020$\bullet$ & 0.681$\pm$0.012$\bullet$ & 0.456$\pm$0.027$\bullet$ & 0.694$\pm$0.020$\bullet$ & 0.220$\pm$0.023$\bullet$	& 0.307$\pm$0.017$\bullet$	& 0.487$\pm$0.018$\bullet$	& 0.473$\pm$0.020$\bullet$	\\
\hline                                                                                                                               				
\multirow{2}{*}{COALA}                                                                                                               				
    & $\mathcal{C}_1$    & 0.665$\pm$0.000$\bullet$ & 0.665$\pm$0.000$\bullet$ & 0.749$\pm$0.000$\bullet$ & 0.415$\pm$0.000$\bullet$ & 0.507$\pm$0.016$\bullet$	& 0.145$\pm$0.013$\bullet$	& 0.473$\pm$0.018$\bullet$	& 0.451$\pm$0.021$\bullet$	\\
    & $\mathcal{C}_2$    & 0.497$\pm$0.000$\bullet$ & 1.000$\pm$0.000$\bullet$ & 0.436$\pm$0.000$\bullet$ & 0.734$\pm$0.000$\bullet$ & 0.216$\pm$0.025$\bullet$	& 0.463$\pm$0.021$\circ$	& 0.461$\pm$0.026$\bullet$	& 0.506$\pm$0.019$\bullet$	\\
\hline                                                                                                                               				
\multirow{2}{*}{De-kmeans}                                                                                                           				
    & $\mathcal{C}_1$    & 0.597$\pm$0.017$\bullet$ & 0.799$\pm$0.018$\bullet$ & 0.655$\pm$0.019$\bullet$ & 0.545$\pm$0.012$\bullet$ & 0.545$\pm$0.016$\bullet$	& 0.142$\pm$0.026$\bullet$	& 0.448$\pm$0.023$\bullet$	& 0.520$\pm$0.015$\circ$	\\
    & $\mathcal{C}_2$    & 0.825$\pm$0.019$\bullet$ & 0.604$\pm$0.021$\bullet$ & 0.576$\pm$0.015$\bullet$ & 0.613$\pm$0.030$\bullet$ & 0.376$\pm$0.028$\bullet$	& 0.123$\pm$0.017$\bullet$	& 0.429$\pm$0.013$\bullet$	& 0.560$\pm$0.022$\circ$	\\
\hline                                                                                                                               				
\multirow{2}{*}{ADFT}                                                                                                                				
    & $\mathcal{C}_1$    & 0.665$\pm$0.000$\bullet$ & 0.665$\pm$0.000$\bullet$ & 0.749$\pm$0.000$\bullet$ & 0.415$\pm$0.000$\bullet$ & 0.507$\pm$0.022$\bullet$	& 0.145$\pm$0.019$\bullet$	& 0.469$\pm$0.022$\bullet$	& 0.567$\pm$0.022$\circ$	\\
    & $\mathcal{C}_2$    & 0.631$\pm$0.014$\bullet$ & 0.782$\pm$0.023$\bullet$ & 0.529$\pm$0.024$\bullet$ & 0.684$\pm$0.017$\bullet$ & 0.419$\pm$0.026$\bullet$	& 0.257$\pm$0.014$\bullet$	& 0.466$\pm$0.019$\bullet$	& 0.520$\pm$0.022$\circ$	\\
\hline                                                                                                                               				
\multirow{2}{*}{MNMF}                                                                                                                				
    & $\mathcal{C}_1$    & 0.665$\pm$0.000$\bullet$ & 0.665$\pm$0.000$\bullet$ & 0.749$\pm$0.000$\bullet$ & 0.415$\pm$0.000$\bullet$ & 0.507$\pm$0.016$\bullet$	& 0.145$\pm$0.027$\bullet$	& 0.464$\pm$0.021$\bullet$	& 0.508$\pm$0.018	\\
    & $\mathcal{C}_2$    & 0.587$\pm$0.012$\bullet$ & 0.727$\pm$0.013$\bullet$ & 0.693$\pm$0.022$\bullet$ & 0.723$\pm$0.015$\bullet$ & 0.435$\pm$0.022$\bullet$	& 0.225$\pm$0.022$\bullet$	& 0.511$\pm$0.015$\bullet$	& 0.507$\pm$0.023	\\

\hline

\multirow{2}{*}{mSC}                                                                                                                				
    & $\mathcal{C}_1$    &0.688$\pm$0.013$\bullet$	&0.411$\pm$0.021$\bullet$	&0.849$\pm$0.019$\bullet$	&0.452$\pm$0.016$\bullet$	&0.685$\pm$0.015$\circ$	&0.284$\pm$0.009$\bullet$	&0.692$\pm$0.014$\circ$	&0.350$\pm$0.011$\bullet$ \\
    & $\mathcal{C}_2$    &0.469$\pm$0.016$\bullet$	&0.729$\pm$0.021$\bullet$	&0.482$\pm$0.010$\bullet$	&0.826$\pm$0.016$\bullet$	&0.362$\pm$0.021$\bullet$	&0.440$\pm$0.012$\bullet$	&0.264$\pm$0.014$\bullet$	&0.545$\pm$0.015$\circ$ \\

\hline

\multirow{2}{*}{NBMC}                                                                                                                				
    & $\mathcal{C}_1$    &0.462$\pm$0.012$\bullet$	&0.763$\pm$0.018$\bullet$	&0.529$\pm$0.003$\bullet$	&0.778$\pm$0.014$\bullet$	&0.817$\pm$0.014$\circ$	&0.361$\pm$0.013$\bullet$	&0.623$\pm$0.016$\circ$	&0.351$\pm$0.012$\bullet$  \\
    & $\mathcal{C}_2$    &0.743$\pm$0.027$\bullet$	&0.554$\pm$0.022$\bullet$	&0.833$\pm$0.021$\bullet$	&0.473$\pm$0.018$\bullet$	&0.459$\pm$0.019$\bullet$	&0.591$\pm$0.018$\circ$	&0.381$\pm$0.021$\bullet$	&0.513$\pm$0.016$\circ$  \\

\hline

\multirow{2}{*}{NBMC-OFV}                                                                                                                				
    & $\mathcal{C}_1$  &0.519$\pm$0.029$\bullet$	&0.836$\pm$0.018$\bullet$	&0.805$\pm$0.011$\bullet$	&0.478$\pm$0.012$\bullet$	&\textbf{0.846$\pm$0.012$\circ$}	&0.386$\pm$0.011$\bullet$	&\textbf{0.855$\pm$0.021$\circ$}	&0.353$\pm$0.018$\bullet$  \\
    & $\mathcal{C}_2$  &0.767$\pm$0.026$\bullet$	&0.602$\pm$0.012$\bullet$	&0.538$\pm$0.012$\bullet$	&0.800$\pm$0.015$\bullet$	&0.475$\pm$0.016$\bullet$	&\textbf{0.612$\pm$0.024$\circ$}	&0.236$\pm$0.016$\bullet$	&\textbf{0.622$\pm$0.025$\circ$}  \\

\hline

\multirow{2}{*}{MISC}
    & $\mathcal{C}_1$    & \textbf{1.000$\pm$0.000} & 0.497$\pm$0.000 & \textbf{1.000$\pm$0.000} & 0.331$\pm$0.000 & 0.654$\pm$0.015 & 0.124$\pm$0.016	& 0.645$\pm$0.024	& 0.456$\pm$0.022	\\
    & $\mathcal{C}_2$    & 0.497$\pm$0.000 & \textbf{1.000$\pm$0.000} & 0.331$\pm$0.000 & \textbf{1.000$\pm$0.000} & 0.255$\pm$0.013 & 0.446$\pm$0.026	& 0.355$\pm$0.018	& 0.505$\pm$0.022	\\
\hline
\hline
\multicolumn{2}{c|}{\multirow{2}[3]{*}{\textbf{NMI}}}
& \multicolumn{2}{c|}{ALOI} & \multicolumn{2}{c|}{DSF} & \multicolumn{2}{c|}{CMUface} & \multicolumn{2}{c}{WebKB} \\
\cline{3-10}
\multicolumn{2}{c|}{} &\multicolumn{1}{c}{Shape} &\multicolumn{1}{c|}{Color} &\multicolumn{1}{c}{Upper-body} &\multicolumn{1}{c|}{Lower-body} &\multicolumn{1}{c}{Identity} &\multicolumn{1}{c|}{Pose} &\multicolumn{1}{c}{University} &\multicolumn{1}{c}{Category} \\
\hline
\multirow{2}{*}{MetaC}
    & $\mathcal{C}_1$    & 0.570$\pm$0.020$\bullet$ & 0.276$\pm$0.020$\bullet$ & 0.820$\pm$0.021$\bullet$ & 0.167$\pm$0.023$\bullet$ & 0.463$\pm$0.028$\bullet$	& 0.141$\pm$0.021$\bullet$	& 0.238$\pm$0.023$\bullet$	& 0.289$\pm$0.028$\bullet$	\\
    & $\mathcal{C}_2$    & 0.442$\pm$0.013$\bullet$ & 0.242$\pm$0.016$\bullet$ & 0.475$\pm$0.016$\bullet$ & 0.474$\pm$0.021$\bullet$ & 0.557$\pm$0.016$\bullet$	& 0.122$\pm$0.019$\bullet$	& 0.205$\pm$0.021$\bullet$	& 0.342$\pm$0.023$\bullet$	\\
\hline                                                                                                                               				
\multirow{2}{*}{MSC}                                                                                                                 				
    & $\mathcal{C}_1$    & 0.661$\pm$0.023$\bullet$ & 0.427$\pm$0.023$\bullet$ & 0.721$\pm$0.018$\bullet$ & 0.441$\pm$0.013$\bullet$ & 0.481$\pm$0.023$\bullet$	& 0.113$\pm$0.024$\bullet$	& 0.234$\pm$0.014$\bullet$	& 0.395$\pm$0.024$\bullet$	\\
    & $\mathcal{C}_2$    & 0.206$\pm$0.030$\bullet$ & 0.606$\pm$0.027$\bullet$ & 0.475$\pm$0.017$\bullet$ & 0.705$\pm$0.023$\bullet$ & 0.286$\pm$0.020$\bullet$	& 0.320$\pm$0.019$\bullet$	& 0.286$\pm$0.022$\bullet$	& 0.417$\pm$0.022$\bullet$	\\
\hline                                                                                                                               				
\multirow{2}{*}{OSC}                                                                                                                 				
    & $\mathcal{C}_1$    & 0.372$\pm$0.019$\bullet$ & 0.472$\pm$0.018$\bullet$ & 0.587$\pm$0.024$\bullet$ & 0.147$\pm$0.021$\bullet$ & 0.463$\pm$0.012$\bullet$	& 0.168$\pm$0.017$\bullet$	& 0.290$\pm$0.015$\bullet$	& 0.391$\pm$0.013$\bullet$	\\
    & $\mathcal{C}_2$    & 0.472$\pm$0.021$\bullet$ & 0.372$\pm$0.015$\bullet$ & 0.090$\pm$0.013$\bullet$ & 0.600$\pm$0.023$\bullet$ & 0.209$\pm$0.017$\bullet$	& 0.294$\pm$0.027$\bullet$	& 0.315$\pm$0.021$\bullet$	& 0.359$\pm$0.024$\bullet$	\\
\hline                                                                                                                               				
\multirow{2}{*}{COALA}                                                                                                               				
    & $\mathcal{C}_1$    & 0.344$\pm$0.000$\bullet$ & 0.344$\pm$0.000$\bullet$ & 0.734$\pm$0.000$\bullet$ & 0.250$\pm$0.013$\bullet$ & 0.628$\pm$0.022$\bullet$	& 0.196$\pm$0.014$\bullet$	& 0.298$\pm$0.023$\bullet$	& 0.368$\pm$0.019$\bullet$	\\
    & $\mathcal{C}_2$    & 0.000$\pm$0.000$\bullet$ & 1.000$\pm$0.000$\bullet$ & 0.314$\pm$0.110$\bullet$ & 0.697$\pm$0.000$\bullet$ & 0.253$\pm$0.009$\bullet$	& 0.389$\pm$0.034$\bullet$	& 0.295$\pm$0.018$\bullet$	& 0.407$\pm$0.024$\bullet$	\\
\hline                                                                                                                              				
\multirow{2}{*}{De-kmeans}                                                                                                           				
    & $\mathcal{C}_1$    & 0.206$\pm$0.021$\bullet$ & 0.606$\pm$0.016$\bullet$ & 0.528$\pm$0.016$\bullet$ & 0.362$\pm$0.022$\bullet$ & 0.590$\pm$0.030$\bullet$	& 0.173$\pm$0.022$\bullet$	& 0.246$\pm$0.018$\bullet$	& 0.448$\pm$0.029$\circ$	\\
    & $\mathcal{C}_2$    & 0.654$\pm$0.030$\bullet$ & 0.211$\pm$0.017$\bullet$ & 0.429$\pm$0.023$\bullet$ & 0.482$\pm$0.010$\bullet$ & 0.559$\pm$0.018$\bullet$	& 0.159$\pm$0.021$\bullet$	& 0.214$\pm$0.021$\bullet$	& 0.475$\pm$0.023$\circ$	\\
\hline                                                                                                                               				
\multirow{2}{*}{ADFT}                                                                                                                				
    & $\mathcal{C}_1$    & 0.344$\pm$0.000$\bullet$ & 0.344$\pm$0.000$\bullet$ & 0.734$\pm$0.000$\bullet$ & 0.250$\pm$0.013$\bullet$ & 0.628$\pm$0.021$\bullet$	& 0.196$\pm$0.011$\bullet$	& 0.291$\pm$0.017$\bullet$	& 0.507$\pm$0.019$\circ$	\\
    & $\mathcal{C}_2$    & 0.272$\pm$0.022$\bullet$ & 0.572$\pm$0.024$\bullet$ & 0.281$\pm$0.016$\bullet$ & 0.559$\pm$0.019$\bullet$  & 0.641$\pm$0.023$\bullet$	& 0.203$\pm$0.014$\bullet$	& 0.302$\pm$0.019$\bullet$	& 0.426$\pm$0.021	\\
\hline                                                                                                                               				
\multirow{2}{*}{MNMF}                                                                                                                				
    & $\mathcal{C}_1$    & 0.344$\pm$0.000$\bullet$ & 0.344$\pm$0.000$\bullet$ & 0.734$\pm$0.000$\bullet$ & 0.250$\pm$0.013$\bullet$ & 0.628$\pm$0.021$\bullet$	& 0.196$\pm$0.021$\bullet$	& 0.292$\pm$0.021$\bullet$	& 0.411$\pm$0.017$\bullet$	\\
    & $\mathcal{C}_2$    & 0.187$\pm$0.011$\bullet$ & 0.587$\pm$0.025$\bullet$ & 0.523$\pm$0.020$\bullet$ & 0.633$\pm$0.014$\bullet$ & 0.554$\pm$0.017$\bullet$	& 0.323$\pm$0.029$\bullet$	& 0.401$\pm$0.029$\bullet$	& 0.269$\pm$0.018$\bullet$	\\

\hline

\multirow{2}{*}{mSC}                                                                                                                				
    & $\mathcal{C}_1$    &0.759$\pm$0.017$\bullet$	&0.319$\pm$0.018$\bullet$	&0.811$\pm$0.017$\bullet$	&0.547$\pm$0.016$\bullet$	&0.754$\pm$0.019$\circ$	&0.252$\pm$0.011$\bullet$	&0.792$\pm$0.014$\circ$	&0.306$\pm$0.012$\bullet$ \\
    & $\mathcal{C}_2$    &0.255$\pm$0.022$\bullet$	&0.698$\pm$0.015$\bullet$	&0.455$\pm$0.019$\bullet$	&0.739$\pm$0.012$\bullet$	&0.244$\pm$0.019$\bullet$	&0.423$\pm$0.015$\bullet$	&0.286$\pm$0.019$\bullet$	&0.511$\pm$0.018$\circ$ \\

\hline

\multirow{2}{*}{NBMC}                                                                                                                				
    & $\mathcal{C}_1$    &0.255$\pm$0.016$\bullet$	&0.785$\pm$0.015$\bullet$	&0.391$\pm$0.016$\bullet$	&0.841$\pm$0.015$\bullet$	&0.797$\pm$0.014$\circ$	&0.385$\pm$0.021$\bullet$	&0.605$\pm$0.009$\circ$	&0.334$\pm$0.016$\bullet$ \\
    & $\mathcal{C}_2$    &0.763$\pm$0.012$\bullet$	&0.224$\pm$0.021$\bullet$	&0.806$\pm$0.012$\bullet$	&0.354$\pm$0.014$\bullet$	&0.451$\pm$0.011$\bullet$	&0.540$\pm$0.015$\circ$	&0.468$\pm$0.017$\bullet$	&0.694$\pm$0.015$\circ$  \\

\hline

\multirow{2}{*}{NBMC-OFV}                                                                                                                				
    & $\mathcal{C}_1$   &0.276$\pm$0.006$\bullet$	&0.860$\pm$0.018$\bullet$	&0.786$\pm$0.018$\bullet$	&0.249$\pm$0.020$\bullet$	&\textbf{0.829$\pm$0.012$\circ$}	&0.352$\pm$0.017$\bullet$	&\textbf{0.857$\pm$0.012$\circ$	} &0.531$\pm$0.018$\bullet$  \\
    & $\mathcal{C}_2$   &0.781$\pm$0.018$\bullet$	&0.341$\pm$0.014$\bullet$	&0.399$\pm$0.014$\bullet$	&0.788$\pm$0.008$\bullet$	&0.35$\pm$0.0151$\bullet$	&\textbf{0.571$\pm$0.019$\circ$}	&0.454$\pm$0.009$\bullet$	&\textbf{0.699$\pm$0.014$\circ$}  \\

\hline

\multirow{2}{*}{MISC}
    & $\mathcal{C}_1$    & \textbf{1.000$\pm$0.000} & 0.000$\pm$0.000 & \textbf{1.000$\pm$0.000} & 0.000$\pm$0.000 & 0.691$\pm$0.019	& 0.221$\pm$0.021	& 0.544$\pm$0.025	& 0.225$\pm$0.016	\\
    & $\mathcal{C}_2$    & 0.000$\pm$0.000 & \textbf{1.000$\pm$0.000} & 0.000$\pm$0.000 & \textbf{1.000$\pm$0.000} & 0.325$\pm$0.023  & 0.501$\pm$0.019	& 0.345$\pm$0.019	& 0.422$\pm$0.015	\\
\hline
\end{tabular}
\label{tab:F1}
\end{table*}
Fig. \ref{fig:aloi_MISC} shows the two clusterings found by MISC on the ALOI dataset: one reveals the subspace corresponding to shape (Fig. \ref{fig:aloi_s1}), and the other subspace corresponding to color (Fig. \ref{fig:aloi_s2}). Similarly, Fig. \ref{fig:dsf_MISC} gives the two clusterings of MISC on the DSF dataset: one reveals the subspace corresponding to the upper-body (Fig. \ref{fig:dsf_s1}), and the other subspace representing the lower-body (Fig. \ref{fig:dsf_s2}). Fig. \ref{fig:Face_MISC} provides two clusterings of MISC on the CMUface dataset: one represents the clustering according to `identity' (Fig. \ref{fig:Face_ID}) and the other according to  `pose' (Fig. \ref{fig:Face_Pose}). All the figures confirm that MISC is capable of finding meaningful clusterings embedded  in the respective subspaces.

MISC gives the best results across both evaluation metrics on each view for ALOI and DSF. Although the competitive algorithms can also find two different clusterings on these two datasets, the corresponding F1 and NMI values are smaller (by at least 20\%) than those of MISC. The reason is that MISC first uses ISA to convert the full feature space into two independent subspaces, and then clusters the data in each subspace. In contrast, De-$k$means and MNMF find  two clusterings in the full feature space, and don't perform well when the actual clusterings are embedded in subspaces. In addition, although ADFT and OSC do explore the second clustering with respect to a feature weighted subspace or a feature-transformed subspace, this clustering is still affected by the reference one, which is computed in the full-space. In contrast, the second clustering explored by MISC is independent from the first one, and  has a meaningful interpretation.

MISC does not perfectly identify the two given clusterings for the CMUface and WebKB datasets. Nevertheless, it can still distinguish the two different views on each dataset. It's possible that these different views embedded in subspaces share some common features and are not completely independent; as such, the two subspaces found by MISC do not quite correspond to the original views. The other methods (De-$k$means, ADFT, and MNMF) cannot well identify the two views, because both $\mathcal{C}_1$ and $\mathcal{C}_2$ are close to the `identity' clustering and far away from the `pose' one. Compared to MISC, De-$k$means finds multiple clusterings in the full space; as such, it cannot discover  clusters embedded in subspaces. MNMF, ADFT, and OSC  find multiple clusterings sequentially, thus subsequent ones depend on the formerly found ones. NBMC-OFV achieves the best results on CMUface and WebKB. The reason is that NBMC-OFV can discover multiple partially overlapping views, whereas the other algorithms can't.

In summary, the advantages of MISC  can be attributed to the explored multiple independent subspaces and to the kernel graph regularized semi-nonnegative matrix factorization, which contribute to the finding of low-redundant clusterings of high-quality.

\section{Conclusion}
In this paper, we study how to find multiple clusterings from data, and present an approach called MISC. MISC assumes that diverse clusterings may be embedded in different subspaces. It first uses independent component analysis to explore statistical independent subspaces, and it determines the number of subspaces and the number of clusters in each subspace. Next, it introduces a kernel graph regularized semi-nonnegative matrix factorization method to find linear and non-linear separable clusters in the subspaces.  Experimental results on synthetic and real-world data demonstrate that MISC can identify meaningful alternative clusterings, and it also outperforms state-of-the-art multiple clustering methods. In the future, we plan to investigate solutions to find alternative clusterings embedded in overlapping subspaces. The code for MISC is available at {http://mlda.swu.edu.cn/codes.php?name=MISC.}

\section{Acknowledgments.}
This work is supported by NSFC (61872300, 61741217, 61873214, and 61871020), NSF of CQ CSTC (cstc2018jcyjAX0228,  cstc2016jcyjA0351, and CSTC2016SHMSZX0824), the Open Research Project of Hubei Key Laboratory of Intelligent Geo-Information Processing (KLIGIP-2017A05), and the National Science and Technology Support Program (2015BAK41B04).


\bibliographystyle{aaai}
\bibliography{AAAI-WangX}

\end{document}